\documentclass{NLE}
\usepackage{savesym}
\savesymbol{checkmark}
\savesymbol{gtrsim}
\savesymbol{lesssim}
\usepackage{pdflscape}
\usepackage{dingbat} 
\usepackage{adjustbox}
\usepackage{amssymb}
\usepackage{multirow}

\usepackage[colorlinks=true,linkcolor=blue,citecolor=blue,urlcolor=black,bookmarksopen=true,bookmarksopenlevel=3,bookmarksdepth=3]{hyperref}
\usepackage[numbered]{bookmark}
\bookmarksetup{startatroot} 

\usepackage{subcaption}
\usepackage{multicol}
\usepackage{natbib}
\usepackage[]{todonotes}
\ifpdf%
\usepackage{epstopdf}%
\else%
\fi

\begin{document}
\label{firstpage}

\lefttitle{Gamified Crowdsourcing for Idiom Corpora Construction}

\papertitle{Article}

\jnlPage{1}{00}
\jnlDoiYr{2021}
\doival{10.1017/xxxxx}

\title{Gamified Crowdsourcing for Idiom Corpora Construction}

\begin{authgrp}
\author{Gülşen Eryiğit$^{1,2,*}$}
\author{Ali Şentaş$^1$}
\author{Johanna Monti$^3$}
\affiliation{$^1$NLP Research Group, Faculty of Computer\&Informatics, Istanbul Technical University, Istanbul, Turkey}
\affiliation{$^2$Department of Artificial Intelligence \& Data Engineering, Istanbul Technical University, Istanbul, Turkey}
\affiliation{$^3$UNIOR NLP Research Group - Department of Literary, Linguistic and Comparative Studies, University of Naples L’Orientale}
\affiliation{*Corresponding author. Email:gulsen.cebiroglu@itu.edu.tr}
\end{authgrp}

\history{(Received 2021; revised xx xxx xxx; accepted xx xxx xxx)}

\begin{abstract}
Learning idiomatic expressions is seen as one of the most challenging stages in second language learning because of their unpredictable meaning. A similar situation holds for their identification within natural language processing applications such as machine translation and parsing. 
The lack of high-quality usage samples exacerbates this challenge not only for humans but also for artificial intelligence systems. 
This article introduces a gamified crowdsourcing approach for collecting language learning materials for idiomatic expressions; a messaging bot is designed as an asynchronous multiplayer game for native speakers who compete with each other while providing idiomatic and nonidiomatic usage examples and rating other players' entries. 
As opposed to classical crowdprocessing annotation efforts in the field, for the first time in the literature, a crowdcreating \& crowdrating approach is implemented and tested for idiom corpora construction. 
The approach is language independent and evaluated on two languages in comparison to traditional data preparation techniques in the field. 
The reaction of the crowd is monitored under different motivational means (namely, gamification affordances and monetary rewards).
The results reveal that the proposed approach is powerful in collecting the targeted materials, and although being an explicit crowdsourcing approach, it is found entertaining and useful by the crowd. The approach has been shown to have the potential to speed up the construction of idiom corpora for different natural languages to be used as second language learning material, training data for supervised idiom identification systems, or samples for lexicographic studies.
\end{abstract}
\maketitle
\section{Introduction} An idiom is usually defined as a group of words established by usage as having a idiosyncratic meaning not deducible from those of the individual words forming that idiom. It is possible to encounter cases where the idiom's constituents come together with or without forming that special meaning. This ambiguous situation poses a significant challenge for both foreign language learners and artificial intelligence (AI) systems since it requires a deep semantic understanding of the language.

Idiomatic control has been seen as a measure of proficiency in a language both for humans and AI systems.
The task is usually referred to as idiom identification or idiom recognition in natural language processing (NLP) studies and is defined as understanding/classifying the idiomatic (i.e., figurative) or nonidiomatic  usage of a group of words (i.e., either with the literal meaning arising from their cooccurrence or by their separate usage). For the words \{``hold'',``one's'', ``tongue''\}, two such usage examples are provided below:\\
 ``Out of sheer curiosity I \textit{held} \textit{my} \textit{tongue}, and waited.'' (idiomatic) (meaning ``stop talking'')\\
``One of the things that they teach in first aid classes is that a victim having a seizure can swallow his tongue, and you should \textit{hold} \textit{his} \textit{tongue} down.'' (nonidiomatic) 


Learning idiomatic expressions is seen as one of the most challenging stages in second language learning because of their unpredictable meaning. 
Several studies have discussed efficient ways of teaching idioms to second language (L2) learners \citep{vasiljevic2015teaching,teachingMWE2017}, and obviously, both computers and humans need high-quality usage samples exemplifying the idiom usage scenarios and patterns. When do some words occurring within the same sentence form a special meaning together? Can the components of an idiom undergo different morphological inflections? If so, is it possible to inflect them in any way or do they have particular limitations? May other words intervene between the components of an idiom? If so, could these be of any word type or are there any limitations? Although it may be possible to deduct some rules (see Appendix for an example) defining some specific idioms, unfortunately, creating a knowledge base that provides such detailed definitions or enough samples to deduct answers to these questions is a very labor-intensive and expensive process, which could only be conducted by native speakers. 
Yet, these knowledge bases are crucial for foreign language learning since there is not enough time or input for students to implicitly acquire idiom structures in the target language.

Due to the mentioned difficulties, there exist very few studies that introduce an idiom corpus (providing idiomatic and nonidiomatic examples), and these are available only for a couple of languages and a limited number of idioms: \citet{birke2006clustering,cook2008vnc} for 25 and 53 English idioms respectively, and \citet{hashimoto2009compilation} for 146 Japanese idioms. 
Similarly, high coverage idiom lexicons either do not exist for every language or contain only a couple of idiomatic usage samples, which is insufficient to answer the above questions. 
Examples of use were considered as musthave features of an idiom dictionary app in \citet{caruso2019can} that tested a dictionary mockup for the Italian language with Chinese students.
On the other hand, it may be seen that foreign language learning communities are trying to fill this resource gap by creating/joining online groups or forums to share idiom examples\footnote{Some examples include 
\url{https://t.me/Idiomsland} for English idioms with 65K subscribers, \url{https://t.me/Deutschpersich} for German idioms with 3.4K subscribers, \url{https://t.me/deyimler} for Turkish Idioms with 2.7K subscribers, \url{https://t.me/Learn_Idioms} for French idioms with 2.5 subscribers. The last three are messaging groups providing idiom examples and their translations in Arabic.}. Obviously, the necessity for idiom corpora applies to all natural languages and we need an innovative mechanism to speed up the creation process of such corpora by ensuring the generally accepted quality standards in language resource creation.

Gamified crowdsourcing is a rapidly increasing trend, and researchers explore creative methods of use in different domains
\citep{morschheuser2017gamified,morschheuser2019gamification, MURILLOZAMORANO2020100645}. 
The use of gamified crowdsourcing for idiom corpora construction has the potential to provide solutions to the above-mentioned problems, as well as to the unbalanced distributions of idiomatic and nonidiomatic samples, and the data scarcity problem encountered in traditional methods.
This article proposes a gamified crowdsourcing approach for idiom corpora construction where 
the crowd is actively taking a role in creating and annotating the language resource and rating annotations.
The approach is experimented on two languages (Turkish and Italian) and evaluated in comparison to traditional data preparation techniques in the field. The results reveal that the approach is powerful in collecting the targeted materials, and although being an explicit crowdsourcing approach, it is found entertaining and useful by the crowd. The approach has been shown to have the potential to speed up the construction of idiom corpora for different natural languages, to be used as
second language learning material, training data for supervised idiom identification systems, or samples for lexicographic studies.

The article is structured as follows: 
${\mathsection}$2 provides a background and the related work, ${\mathsection}$3 describes the game design, ${\mathsection}$4 provides analyses, and  ${\mathsection}$5  the conclusion.
\section{Background \& Related Work}\label{sec:background} Several studies investigate idioms from a cognitive science perspective: \citet{kaschak2006idiomatic} constructed artificial grammars that contained idiomatic and ``core'' (nonidiomatic) grammatical rules and examined learners’ ability to learn the rules from the two types of constructions. The findings suggested that learning was impaired by idiomaticity, counter to the conclusion of \citet{sprenger2006lexical} that structural generalizations from idioms and nonidioms are similar in strength.
\citet{konopka2009lexical} investigate idiomatic and non-idiomatic English phrasal verbs and states that
despite differences in idiomaticity and structural flexibility, both types of
phrasal verbs induced structural generalizations and differed little in their ability to do so.

We may examine the traditional approaches which focus on idiom annotation in two main parts:
first, the studies focusing solely on idiom corpus construction, and
second the studies on general multiword expressions' (MWEs) annotations also including idioms. 
Both approaches have their own drawbacks, and exploration of different data curation strategies in this area is crucial for any natural language, but especially for morphologically rich and low resource languages (MRLs and LRLs).

The studies focusing solely on idiom corpus construction \citep{birke2006clustering,cook2008vnc,hashimoto2009compilation} first retrieve sentences from a text source according to some keywords from the target group of words (i.e., target idiom's constituents) and then annotate them as idiomatic or nonidiomatic samples.
The retrieval process is not as straightforward as one might think since the keywords should cover all possible inflected forms of the words in focus (e.g. keyword ``go'' could not retrieve its inflected form ``went''), especially for MRLs where words may appear under hundreds of different surface forms. 
The solution to this may be lemmatization of the corpus and searching with lemmas, but this will not work in cases where the data source is pre-indexed and only available via a search engine interface such as the internet.
This first approach may also lead to unexpected results on the class distributions. For example, \citet{hashimoto2009compilation} states that examples were annotated for each idiom, regardless of the proportion of idioms and literal phrases, until the total number of examples for each idiom reached 1000, which is sometimes not reachable due to data unavailability.

Idioms are seen as a subcategory\footnote{In this article, differing from \citet{mweprocessingsurvey}, which list subcategories of MWEs, we use the term ``idiom'' for all types of MWEs carrying an idiomatic meaning including phrasal verbs in some languages.} of multiword expressions (MWEs) which have been subject to many initiatives in recent years such as Parseme EU COST Action, MWE-LEX workshop series, ACL special interest group SIGLEX-MWE.
Traditional methods for creating MWE corpora \citep{schneider2014comprehensive,Vincze11,losnegaard2016parseme,Savarytv} generally rely on manually annotating MWEs on previously collected text corpora (news articles most of the time and sometimes books), this time without being retrieved with any specific keywords. 
However, the scarcity of MWEs (especially idioms) in text have presented obstacles to corpus-based studies and NLP systems addressing these \citep{schneider2014comprehensive}. 
In this approach, only idiomatic examples are annotated. One may think that all the remaining sentences containing idiom's components are nonidiomatic samples. 
However, in this approach, human annotators are prone to overlook especially those MWE components that are not juxtaposed within a sentence.
\citet{bontcheva2017crowdsourcing} states that annotating one named entity (another sub-category of MWEs) type at a time as a crowdsourcing task is a better approach than trying to annotate all entity types at the same time.
Similar to \citet{bontcheva2017crowdsourcing}, our approach achieves the goal of collecting quality and creative samples by focusing the crowd's attention on a single idiom at a time. Crowdsourcing MWE annotations has been rarely studied 
\citep{kato2018construction,fort2018fingers,fort-etal-2020-rigor} and these were crowdprocessing\footnote{``Crowdprocessing approaches rely on the crowd to perform large quantities of homogeneous tasks. Identical contributions are
a quality attribute of the work's validity. The value is derived directly from each isolated contribution (non-emergent)'' \citep{morschheuser2017gamified}.} efforts.




Crowdsourcing \citep{howe2006rise} is a technique used in many linguistic data collection tasks \citep{mitrovic2013crowdsourcing}. 
Crowdsourcing systems are categorized under four main categories: crowdprocessing, crowdsolving, crowdrating, and crowdcreating \citep{GEIGER20143,PRPIC201577,morschheuser2017gamified}. 
While ``Crowdcreating solutions seek to create comprehensive (emergent) artifacts based on a variety of heterogeneous contributions'', `` Crowdrating systems commonly seek to harness the so-called wisdom of crowds to perform collective assessments or predictions'' \citep{morschheuser2017gamified}. The use of these two later types of crowdsourcing together has a high potential to provide solutions to the above-mentioned problems for idiom corpora construction.

One platform researchers often use for crowdsourcing tasks is \textit{Amazon Mechanical Turk (MTurk)}\footnote{https://www.mturk.com}. \cite{snow2008cheap} used it for linguistics tasks such as word similarity, textual entailment, temporal ordering, and word sense disambiguation. \cite{lawson2010annotating} used MTurk to build an annotated NER corpus from emails. \cite{akkaya2010amazon} used the platform to gather word-sense disambiguation data. The platform proved especially cost-efficient in the highly human labor-intensive task of word-sense disambiguation \citep{akkaya2010amazon, rumshisky2012word}. Growing popularity also came with criticism for the platform as well \citep{fort2011amazon}.

MTurk platform uses monetary compensation as an incentive to complete the tasks. Another way of utilizing the crowd for microtasks is gamification, which, as an alternative to monetary compensation utilizes game elements such as points, achievements, and leaderboards. \cite{von2006games} pioneered these types of systems and called them Games with a Purpose (GWAP)~\citep{von2006games}. ESPGame \citep{von2004labeling} can be considered as one of the first GWAPs. It's designed as a game where users were labeling images from the web while playing a Taboo™ like game against each other. The authors later developed another GWAP, Verbosity~\citep{von2006verbosity}, this time for collecting common-sense facts in a similar game setting. GWAPs are popularized in the NLP field by early initiatives such as \textit{1001 Paraphrases}~\citep{chklovski2005collecting}, \textit{Phrase Detectives} \citep{chamberlain2008addressing}, \textit{JeuxDeMots} \citep{artignan2009multiscale}, \textit{Dr. Detective} \citep{dumitrache2013dr}.
\textit{RigorMortis} \citep{fort2018fingers,fort-etal-2020-rigor} gamifies the traditional MWE annotation process described above.

Gamified crowdsourcing is a rapidly increasing trend, and researchers explore creative methods of use in different domains \citep{morschheuser2017gamified,morschheuser2019gamification, MURILLOZAMORANO2020100645}. \citet{morschheuser2017gamified} introduce a conceptual framework of gamified crowdsourcing systems according to which the motivation of the crowd may be provided by either gamification affordances (such as leaderboards, points, and badges) or additional incentives (such as monetary rewards, prizes). In our study, we examine both of these motivation channels and report their impact. According to \citet{morschheuser2017gamified}, ``one major challenge in motivating people to participate is to design a crowdsourcing system that promotes and enables the formation of positive motivations towards crowdsourcing work and fits the type of the activity.'' Our approach to gamified crowdsourcing for idiom corpora construction relies on crowdcreating and crowdrating. We both value the creativity and systematic contributions of the crowd. As explained above, since it is not easy to retrieve samples from available resources, we expect our users to be creative in providing high-quality samples. 

\citet{MORSCHHEUSER2018219} state that users increasingly expect the software to be not only useful but also enjoyable to use, and a gamified software requires an in-depth understanding of motivational psychology and requires multidisciplinary knowledge. In our case, these multidisciplines include language education, computational linguistics, natural language processing, and gamification.
To shed light to a successful design of gamified software, the above-mentioned study divides the engineering of gamified software into 7 main phases and mention 13 design principles (from now on depicted as DP\#n where n holds for the design principle number) adopted by experts.
These 7 main phases are project preparation, analysis, ideation, design, implementation, evaluation, and monitoring phases. The following sections provide the details of our gamification approach by relating the stages to these main design phases and principles.




\section{Game Design} The aim while designing the software was to create an enjoyable and cooperative environment that would motivate the volunteers to help the research studies.
The game is designed to collect usage samples for idioms of which the words of the idiom may also commonly be used in their literal meanings within a sentence. 
An iterative design process has been adopted. 
After the first ideation, design, and prototype implementation phases, the prototype was shared with the stakeholders (see ${\mathsection}$Acknowledgments) (as stated in DP\#7) and the design has been improved accordingly.

A messaging bot (named ``Dodiom''\footnote{A language-agnostic name has been given for the game to be generalized to every language.}) is designed as an asynchronous multiplayer game\footnote{Asynchronous multiplayer games enable players to take their turns at a time that suits them; i.e., the users do not need to be in the game simultaneously.} for native speakers who compete with each other while providing idiomatic and nonidiomatic usage examples and rating other players' entries. 
The game is an explicit crowdsourcing game and players are informed from the very beginning that they are helping to create a public data source by playing this game\footnote{In addition to the reporting and banning mechanism (DP\#10), we also shared a consent agreement message in the welcome screen and the announcements in line with DP\#12 related to legal\&ethical constraints.}.

The story of the game is based on a bird named Dodo (the persona of the bot) trying to learn a foreign language and having difficulty learning idioms in that language. Players try to teach Dodo the idioms in that language by providing examples.
Dodiom has been developed as an opensource project (available on Github\footnote{\url{https://github.com/Dodiom/dodiom}}) with the focus on being easily adapted to different languages.
All the interaction messages are localized and shown to the users in the related language; localizations are currently available  for English, Italian, and Turkish languages.

\subsection{Main Interactions and Gameplay}

Dodo learns a new idiom every day.
The idioms to be played each day are selected by moderators according to their tendency 
to be used with their literal meaning.
For each idiom, players have a predetermined time frame to submit their samples and reviews so they can play at their own pace.
Since the bot may send notifications via the messaging platform in use, the time frame is determined as between 11 a.m. and 11 p.m.\footnote{Different time frames have been tried in the iterative development cycle (DP\#4) and it has been decided that this time frame is the most suitable.}

\begin{figure}[htb]
	\centering

\begin{subfigure}[t]{0.45\textwidth}
    \centering
    \includegraphics[width=\textwidth]{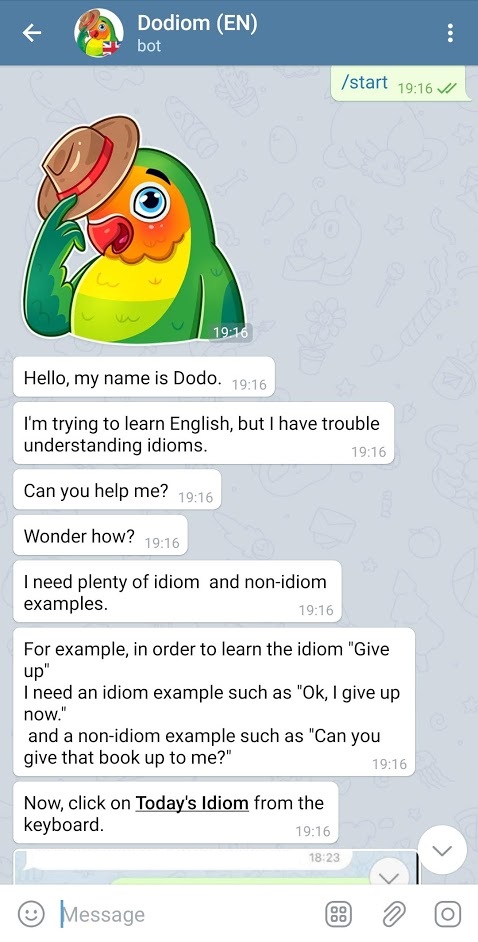}
    \caption{Dodo greeting the player, describing the game, and showing the next steps}
    \label{fig:dodo_greeting}
\end{subfigure}
\hspace{1cm}
\begin{subfigure}[t]{0.45\textwidth}
    \centering
    \includegraphics[width=\textwidth]{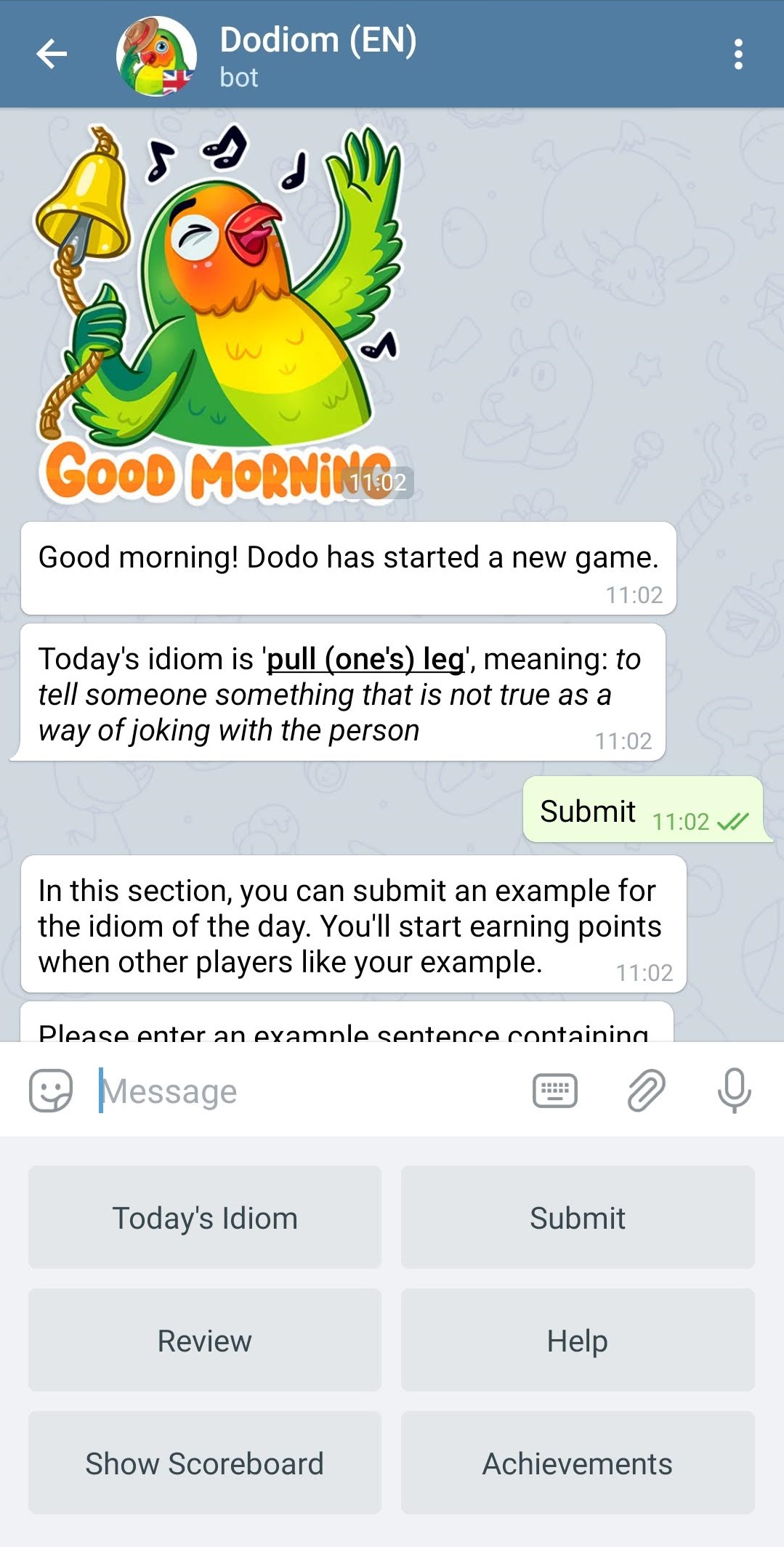}
    \caption{Main Menu showing the currently available options}
    \label{fig:main_menu}
\end{subfigure}
\caption{Dodiom welcome and menu screens.}
	\label{fig:dodo1}
\end{figure}
 
When the users connect to the bot for the first time, they are greeted with Dodo explaining to them what the game is about and teaching them how to play in a step-by-step manner (Figure~\ref{fig:dodo_greeting}). 
This pre-game tutorial and the simplicity of the game proved useful as most of the players were able to play the game in a matter of minutes and provided high-quality examples. All the game messages are studied very carefully to achieve this goal and ensure that the crowd unfamiliar with AI or linguistics understands the task easily.
Random tips for the game are also shared with the players right after they submit their examples. This approach is similar to video games where tips about the game are shown to players on loading screens and/or menus.

Figure \ref{fig:main_menu} shows the main menu, from where the players can access various modes of the game. 

\textbf{Today's Idiom} tells the player what that day's chosen idiom is, players can then submit usage examples for said idioms to get more points.

\textbf{Submit} allows players to submit new usage examples. When clicked, Dodo asks the player to input the example sentence and when the player sends one, the sentence is checked if it contains the words (i.e., the lemmas of the words) that appear in the idiom. If so, Dodo then asks whether these words form an idiom in the given sentence or not. The players are awarded each time other players like their examples so they are incentivized to enter multiple high-quality submissions.

\begin{figure}
\begin{minipage}{.50\textwidth}
 	\begin{subfigure}{\linewidth}
    \centering
    \includegraphics[width=.7\linewidth]{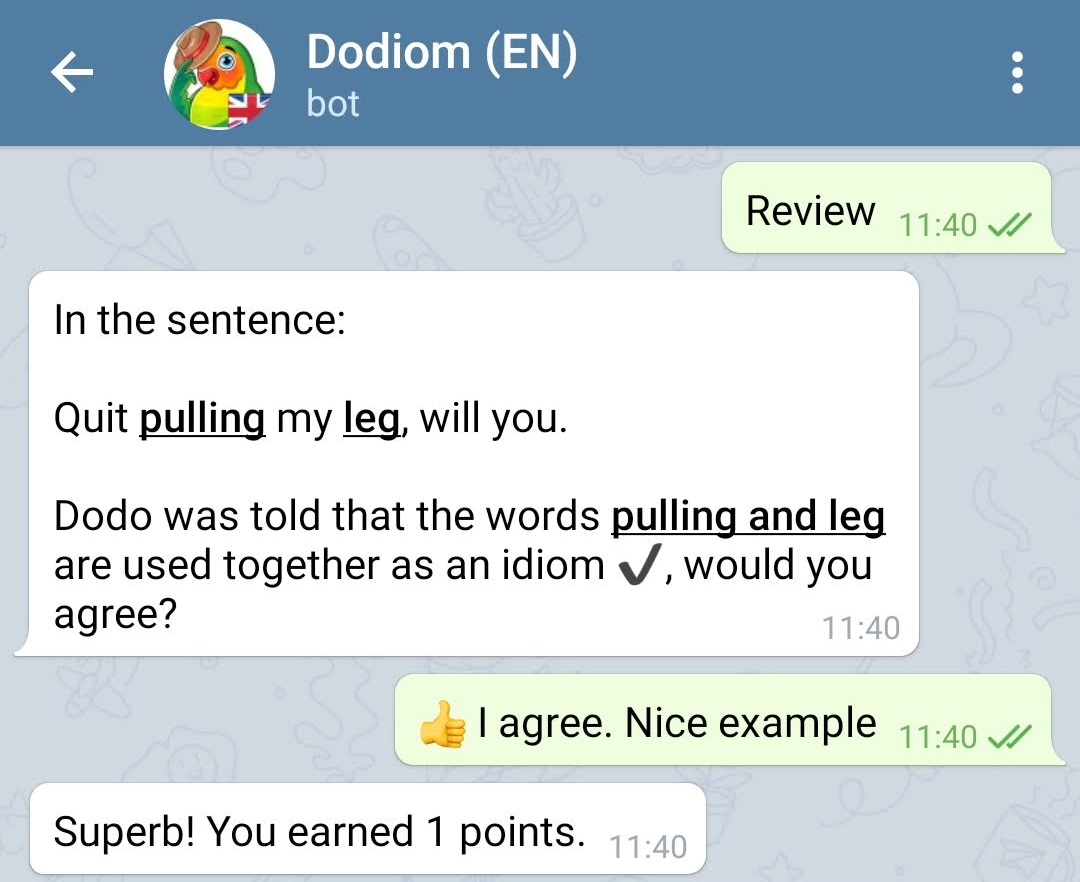}
    \caption{Review interaction}
    \label{fig:review}
\end{subfigure}	\\[1ex]
  \begin{subfigure}{\linewidth}
    \centering
    \includegraphics[width=.7\linewidth]{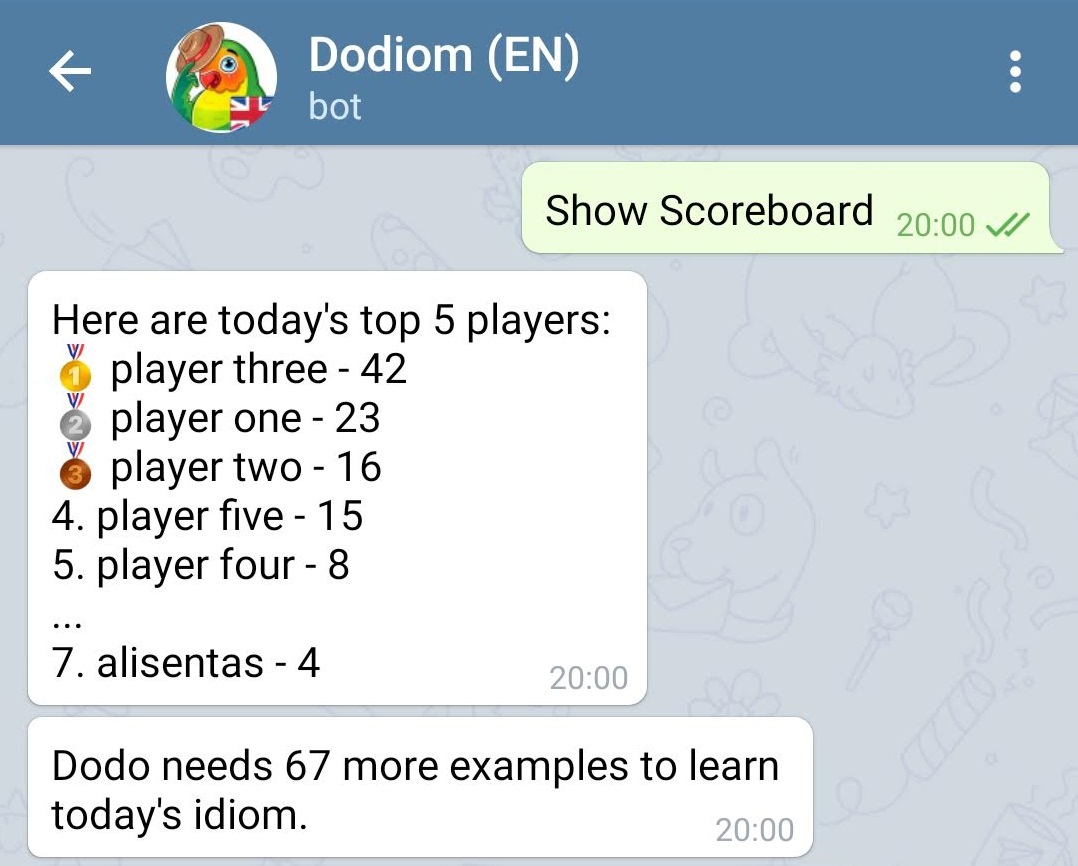}
    \caption{Leaderboard interaction}
    \label{fig:scoreboard}
  \end{subfigure}
\end{minipage}%
\begin{minipage}{.50\textwidth}
  \begin{subfigure}{\linewidth}
    \centering
    \includegraphics[width=.7\linewidth]{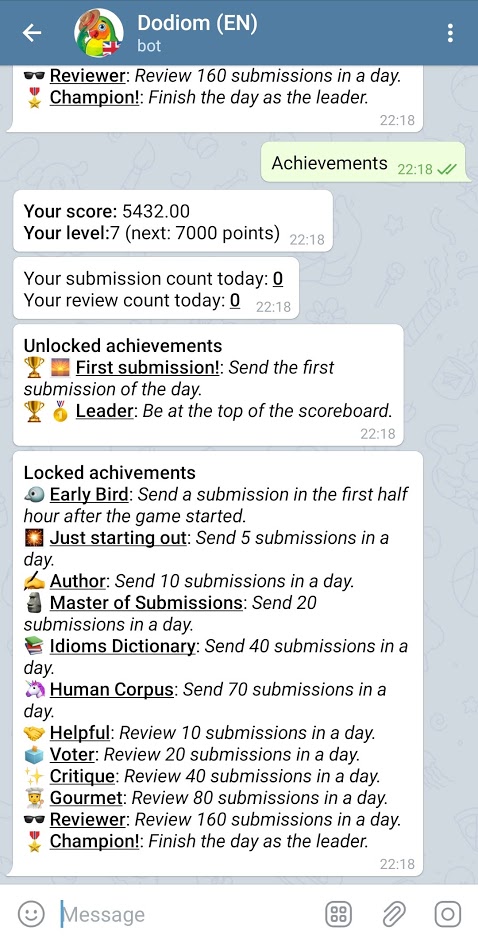}
      \caption{User's score and achievements}
    \label{fig:achievements}
  \end{subfigure}
\end{minipage}%

\caption{Some interaction screens}
\label{fig:test}
\end{figure}

\textbf{Review} allows players to review submissions sent by other players. 
Dodo shows players examples of other players one at a time together with their annotations (i.e., idiom or not), and asks its approval.
Users are awarded points for each submission they review so they are also incentivized to review. The exact scoring and incentivization system will be explained in ${\mathsection}$\ref{sec:gamification}.
Figure \ref{fig:review} shows a simple interaction between Dodo and a user, where Dodo asks whether or not the words \textit{pulling} and \textit{leg} (automatically underlined by the system) in the sentence \textit{``Quit pulling my leg, will you''}  are used idiomatically. 
The user responds with acknowledgment or dislike and then Dodo thanks the user for his/her contribution. Users can also report the examples which don't fit the general guidelines (e.g., vulgar language, improper usage of the platform) for the submissions to be later reviewed by moderators. The moderators can flag the submissions and ban the users from the game depending on the submission. 
Submissions with fewer reviews are shown to the players first (i.e., before the samples that were reviewed previously) so that each submission can receive approximately the same number of reviews.

\textbf{Help} shows the help message, which is a more compact version of the pre-game tutorial.

\textbf{Show Scoreboard} displays the current state of the leaderboard which is updated every time a player is awarded any points. As seen in Figure \ref{fig:scoreboard}, the scoreboard displays the first five players' and the current player's scores. The scoreboard is reset every day for each idiom. Additionally, 100 submissions are set as a soft target for the crowd and a message stating the number of submissions remaining to reach this goal is shown below the scoreboard. The message no longer appears when the target is reached.


\textbf{Achievements} option shows the score, level, and locked/unlocked achievements of the user. An example can be seen in Figure \ref{fig:achievements} where many achievements are designed to gamify the process and reward players for specific actions such as \textit{Early Bird} achievement for early submissions and \textit{Author} for sending 10 submissions in a given day. Whenever an achievement is obtained, the user is notified with a message and an exciting figure (similar to the ones in Figure~\ref{fig:notification}).

\subsection{Gamification affordances \& Additional incentives}
\label{sec:gamification}
Dodiom uses both gamification affordances and additional incentives \citep{morschheuser2017gamified} for the motivation of its crowd.
Before the decision of the final design, we have tested with several scoring systems with and without additional incentives.
This section provides the detailed form of the final scoring system together with previous attempts, gamification affordances, and additional incentives.


\begin{figure}[htb]
	\centering

\begin{subfigure}[t]{0.425\textwidth}
    \centering 
    \includegraphics[width=\textwidth]{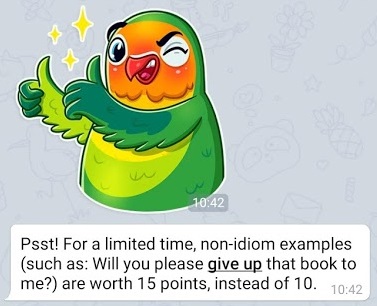}
		 \caption{Score increase}
    \label{fig:notification1}
  \end{subfigure}	
	\hspace{1cm}
	\begin{subfigure}[t]{0.40\textwidth}
    \centering 
 \includegraphics[width=\textwidth]{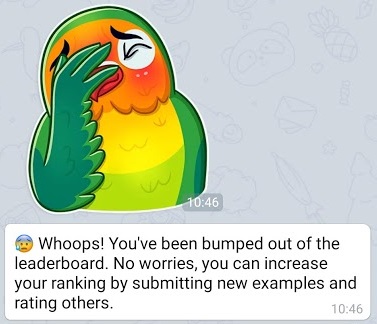}
 \caption{Falling back from the top 5}
    \label{fig:notification2}
\end{subfigure}
    \caption{Notification samples}
    \label{fig:notification}
\end{figure}


The philosophy of the game is based on collecting valuable samples that illustrate the different ways of use and make it possible to make inferences that define how to use a specific idiom (such as the ones in the Appendix).
The samples to be collected are categorized into 4 main types given below. For the sake of game simplicity, this categorization is not explicitly described to the users but is only used for background evaluations.

\begin{itemize}
	\item A-type samples: Idiomatic samples in which the constituent words are used side by side (juxtaposed) (ex: ``Please \textbf{hold} \textbf{your} \textbf{tongue} and wait.''),
	\item B-type samples: Idiomatic samples in which the constituent words are separated by some other words, which is a more common phenomenon in free-word order languages\footnote{In free word order languages, the syntactic information is mostly carried at word level due to affixes thus the words may freely change their position within the sentence without affecting the meaning.} (ex: ``Please \textbf{hold} \textbf{your} breath and \textbf{tongue} and wait for the exciting announcements.''),
	\item C-type samples: Nonidiomatic samples in which the constituent words are used side by side
 	(ex: ``Use sterile tongue depressor to \textbf{hold} \textbf{patient's} \textbf{tongue} down.''),
	\item D-type samples: Nonidiomatic samples in which the constituent words are separated by some other words
	(ex: ``\textbf{Hold} on to \textbf{your} mother \textbf{tongue}.''). 
\end{itemize}

Producing samples from some categories (e.g., B-types and C-types) may not be the most natural form of behavior.
For Turkish, we experimented with different scorings that will motivate our users to produce samples from different categories.
Before settling on the final form of the scoring system, two other systems have been experimented with in the preliminary tests. These include having a fixed set of scores for each type (i.e., 30, 40, 20, and 10 for A-, B-, \mbox{C-,} and D-types respectively). This scoring system caused players to enter submissions for only the B-type to get the most out of a submission and resulted in very few other types of samples.
To fix the major imbalance problem, in another trial, a decay parameter has been added to lower the initial type scores whenever a new submission arrives. 
Unfortunately, this new system had little to no effect on remedying the imbalance and made the game harder to understand for players who couldn't easily figure out the scoring system\footnote{User feedbacks are taken via personal communication on trial runs.}. This latter strategy was also expected to incentivize players to enter submissions early in the game but it didn't work out as planned.

Although being one of the most challenging types for language learners and an important type that we want to collect samples from,
B-type utterances may not be as common as A-type utterances for some idioms and be rare for some languages with fixed word-order.
Similarly producing C-type samples may be difficult for some idioms and overdirecting the crowd to produce more examples of this type can result in unnatural sentences. Thus, game motivations should be chosen carefully. 

We used scoring, notifications, and tips to increase the type variety in the dataset in a meaningful and natural way.
The final scoring system used during the evaluations (presented in the next sections) is as follows:
Each review is worth one point unless it is done in the happy hour during which all reviews are worth two points.
As stated above, after each submission a random tip is shown to the submitter motivating him/her to either review other's entries or to submit samples from either B-type or C-type.
The scores for each type are set to 10 with the only difference of B-type being set to 12.
The system periodically checks the difference between A-type and C-type samples and when this exceeds 15 samples, it increases the scores of the idiomatic or nonidiomatic classes\footnote{That is to say, when \#A-type samples $\geqslant$ \#C-type samples + 15, the scores of C-type and D-type samples are increased by 5, and similarly when \#C-type samples $\geqslant$ \#A-type samples + 15, the scores of A-type and B-type samples are increased by 5.}. The score increase is notified to the crowd via a message (Figure~\ref{fig:notification1} stating either Dodo needs more idiomatic samples or nonidiomatic samples) and remains active until the difference falls below 5 samples. 
As stated above, although for some idioms producing C-type samples may be difficult, 
since the notification message is for calling nonidiomatic samples in general, the crowd is expected to provide both C-type and D-type samples in accordance with the natural balance.


Push notifications are also used to increase player engagement. There are several notifications sent through the game, which are listed below.
The messages are arranged so that an inactive user would only receive a couple of notifications from the game each day; the first three items below are sent to every user whereas the last three are sent only to active users of that day.
\begin{enumerate}
  \item Every morning Dodo sends a good morning message when the game starts and tells the player that day's idiom.
  \item When a category score is changed, a notification is sent to all players (Figure~\ref{fig:notification1}).
  \item A notification is sent to players when review happy hour is started. This event is triggered manually by moderators, and for one hour, points for reviews worth double. This notification also helps to reactivate low-speed play.
  \item When a player's submission is liked by other players, the author of the submission is notified and encouraged to check back the scoreboard. 	Only one message of this type is sent within a limited time to avoid causing too many messages consecutively.
  \item When a player becomes the leader of the scoreboard or enters the first five he/she is congratulated.
  \item When a player loses his/her first position on the leader board or loses his/her place in the first three or five he/she is notified about it and encouraged to get back and send more submissions to take his/her place back. (Figure~\ref{fig:notification2})
\end{enumerate}

We've seen that player engagement increased dramatically when these types of notifications were added (this will be detailed in Section \ref{sec:survey}). 
As additional incentives, we also tested with some monetary rewards given to the best player of each day and investigated the impacts; a 5 Euro online store gift card for Italian, and a 25 Turkish Lira online bookstore gift card for Turkish.

\subsection{Game implementation}

The game is designed as a Telegram bot to make use of Telegram's advanced features (e.g. multi-platform support) which allowed us to focus on the NLP back-end rather than building web-based or mobile versions of the game.
Python-telegram-bot\footnote{\url{https://github.com/python-telegram-bot/python-telegram-bot/}} library is used to communicate with the Telegram servers and to implement the main messaging interface. A PostgreSQL\footnote{\url{https://www.postgresql.org/}} database is used as the data back-end. The ``Love Bird'' Telegram sticker package has been used for the visualization of the selected persona, which can be changed according to the needs (e.g. with a local cultural character). 
For NLP related tasks, NLTK \citep{loper2002nltk} is used for tokenization. Idioms are located in new submissions by tokenizing the submission and checking the lemma of each word whether they match that day's idiom constituents. If all idiom lemmas are found within the submission, the player is asked to choose whether the submission is an idiomatic or nonidiomatic sample. The position of the lemmas determines the type (i.e., one of the for types introduced in Section~\ref{sec:gamification}) of the submission within the system.
NLTK is used for the lemmatization of English, Tint\footnote{A Stanza\citep{qi2020stanza} based tool customized for the Italian language} \citep{palmeromorettitint} for the Italian and Zeyrek\footnote{An NLTK based lemmatizer, customized for the Turkish language, \url{https://zeyrek.readthedocs.io}} for the lemmatization of Turkish\footnote{Stanza is also tested for Turkish, but outputting only a single possible lemma for each word failed in many cases in this language.}.



The game is designed with localization in mind. The localization files  are currently available in English, Italian and Turkish. 
Adaptation to other languages requires: 1. translation of localization files containing game messages (currently 145 interaction messages in total), 2. a list of idioms, and 3. a lemmatizer for the target language. 
We also foresee that there may be need for some language specific enhancements (such as the use of wildcard characters, or words) in the definition of idioms to be categorized under different types.
The game is deployed on Docker\footnote{\url{https://docker.com/}} containers adjusted to each countries time-zone where the game is played. 
In accordance with DP\#4 and DP\#11, an iterative development process has been applied. The designs (specifically the bot's messages, their timings, and frequencies) are tested and improved until they become efficient and promising to reach the goals. 
The system has been monitored and optimized according to the increasing workload.

\section{Analysis \& Discussions}\label{sec:data-analysis}

In accordance with DP\#9, we made a detailed analysis of the collected data set to evaluate the success of the proposed approach for idiom corpora construction, and quantitative and qualitative analysis to evaluate its psychological and behavioral effects on users.
This section first introduces the methodology and participants in $\mathsection$\ref{sec:partic} and then provides an analysis of the collected data in $\mathsection$\ref{sec:an}.
It then gives a comparison with the data collected by using a traditional data annotation approach in $\mathsection$\ref{sec:comp}.
The section then concludes in $\mathsection$\ref{sec:survey} by the analysis of motivational and behavioral outcomes according to some constructs selected from the relevant literature.


\subsection{Methodology \& Participants}
\label{sec:partic}
The game was deployed three times: the first one for preliminary testing with a limited number of users, and then two consecutive 16-day periods open to crowd, for Turkish and Italian separately. 
The first preliminary testing of the game was accomplished on Turkish with nearly 10 people, and yielded to significant improvements
in the game design.
The Italian preliminary tests were accomplished with around 100 people\footnote{Students of the third author and people contacted at EU Researchers Night at Italy.}. 
The game was played between October 13 and December 17, 2020 for Turkish, and between November 8 and December 29, 2020 for Italian.
From now on, the four later periods (excluding the preliminary testing periods), for which we provide data analysis, will be referred to as TrP1, TrP2 for Turkish, ItP1, and ItP2 for Italian. While TrP1 and ItP1 are trials without monetary rewards, TrP2 and ItP2 are with monetary rewards.

The idioms to be played each day were selected by moderators according to their tendency to be used with their literal meaning.
For ItP1 and ItP2, the selection procedure was random from an Italian idiom list\footnote{\url{http://www.impariamoitaliano.com/frasi.htm}}, wherein the later one 4 idioms from ItP1 are replayed for comparison purposes.
Similarly for TrP2, the idioms were randomly selected from an online Turkish idiom list\footnote{\url{https://www.dilbilgisi.net/deyimler-sozlugu/}} again taking two idioms from TrP1 for comparison.
For TrP1, the idioms were selected again with the same selection strategy but this time instead of using an idiom list, the idioms from a previous annotation effort (Parseme multilingual corpus of verbal multiword expressions \citet{Savarytv,ramisch2018edition}) are listed according to their frequencies within the corpus and given to the moderators for the selection.
Table~\ref{tab:TRrun} and Table~\ref{tab:ITRun}, given in the Appendix section, provide the idioms played each day together with daily submission, review statistics and some extra information to be detailed later.

\begin{table}[!htb]
\caption{User Statistics}
\label{tab:user}
\centering
\begin{tabular}{lcc}
\hline
Statistic & Turkish & Italian \\ \hline
Total \# of users who played the game & 255 & 205 \\
 \quad\quad ... for only 1 day & 113 & 93 \\
 \quad\quad ... for 2-3 days & 87 & 61 \\
 \quad\quad ... for 4-7 days & 31 & 32 \\
 \quad\quad ... for \textgreater{}= 7 days & 24 & 19 \\
Total \# of users who filled in the survey: & 25 & 31 \\
\# of days the survey was open: & last 3 days of TrP2 & last 10 days of ItP2 \\
Crowd Type & AI-related people & students, translators \\ \hline
\end{tabular}
\end{table}

For the actual play, the game was announced on LinkedIn and Twitter for both languages at the beginning of each play (viz., TrP1, TrP2, ItP1, ItP2).
For Italian, announcements and daily posts were also shared via Facebook and Instagram.
In total, there were $\sim$25K views and $\sim$400 likes/reshares for Turkish, $\sim$12K views $\sim$400 likes/reshares for Italian.
As mentioned in the previous sections, players are informed from the very beginning that they are helping to create a public data source by
playing this game. It should be noted that many people wanted to join this cooperative effort and shared the announcements from their accounts, which improved the view counts.
For both languages, the announcements of the second 16-day period with monetary reward were also shared within the game itself.
The Turkish crowd influencer (the first author of this article) is from NLP and AI community, and the announcements mostly reached her NLP focused network. On the other hand, the Italian crowd influencer (the last author of this article) is from the computational linguistics community, and the announcements mostly reached students and educators.
There were 255 and 205 players who played the game in total for both periods.
Table \ref{tab:user} provides the detailed user statistics. As may be seen from this table, almost 10 per cent of the players played the game for more than 7 days.
A survey has been shared with the users at the end of TrP2 and ItP2. More than 10 per cent of the players filled in this survey.




\begin{figure}[htb]
	\centering
	\begin{subfigure}[b]{0.48\textwidth}
	    \includegraphics[width=\textwidth]{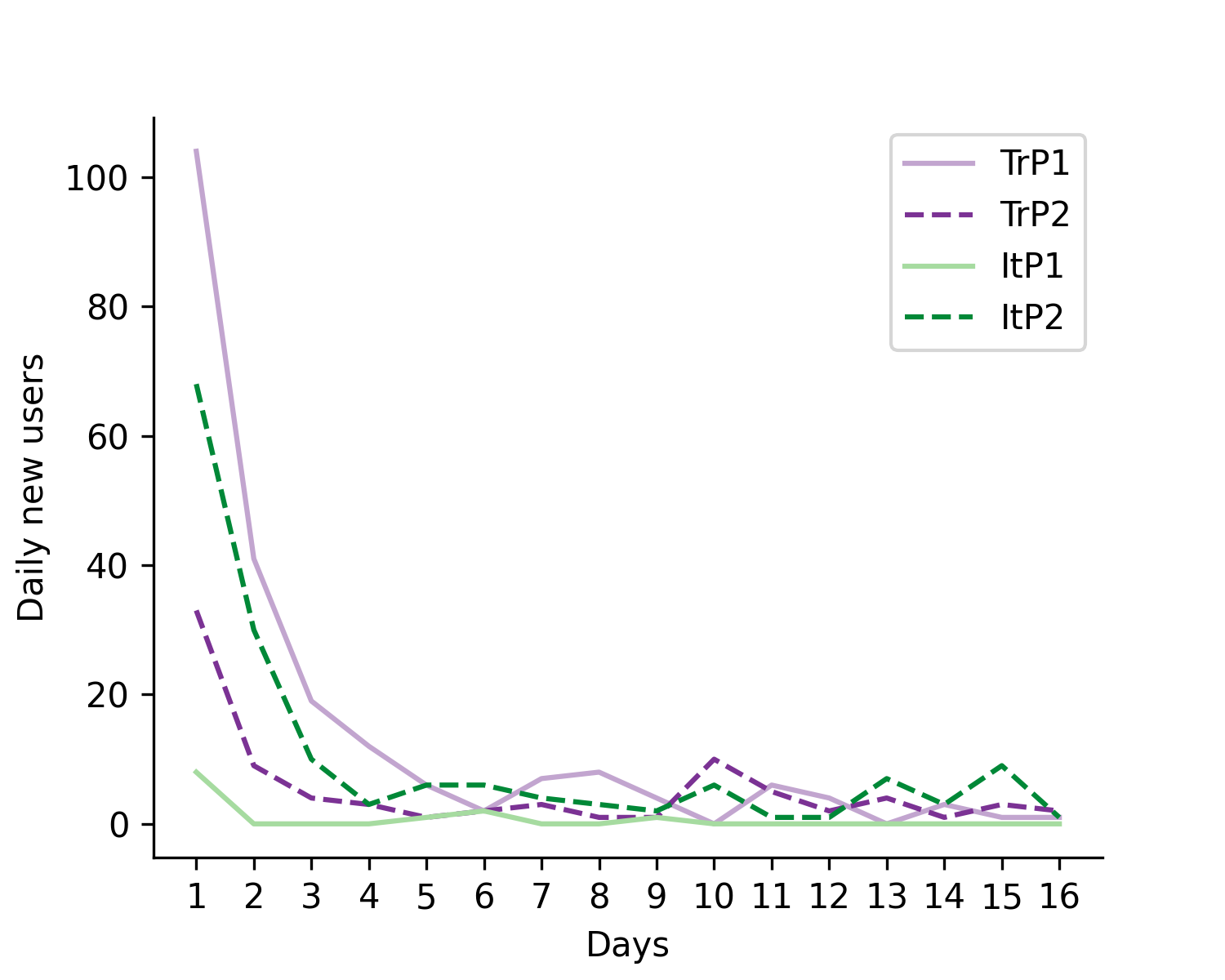}
	    \caption{Daily new player count}
	    \label{fig:daily_new_players}
	\end{subfigure}
	\begin{subfigure}[b]{0.48\textwidth}
	    \includegraphics[width=\textwidth]{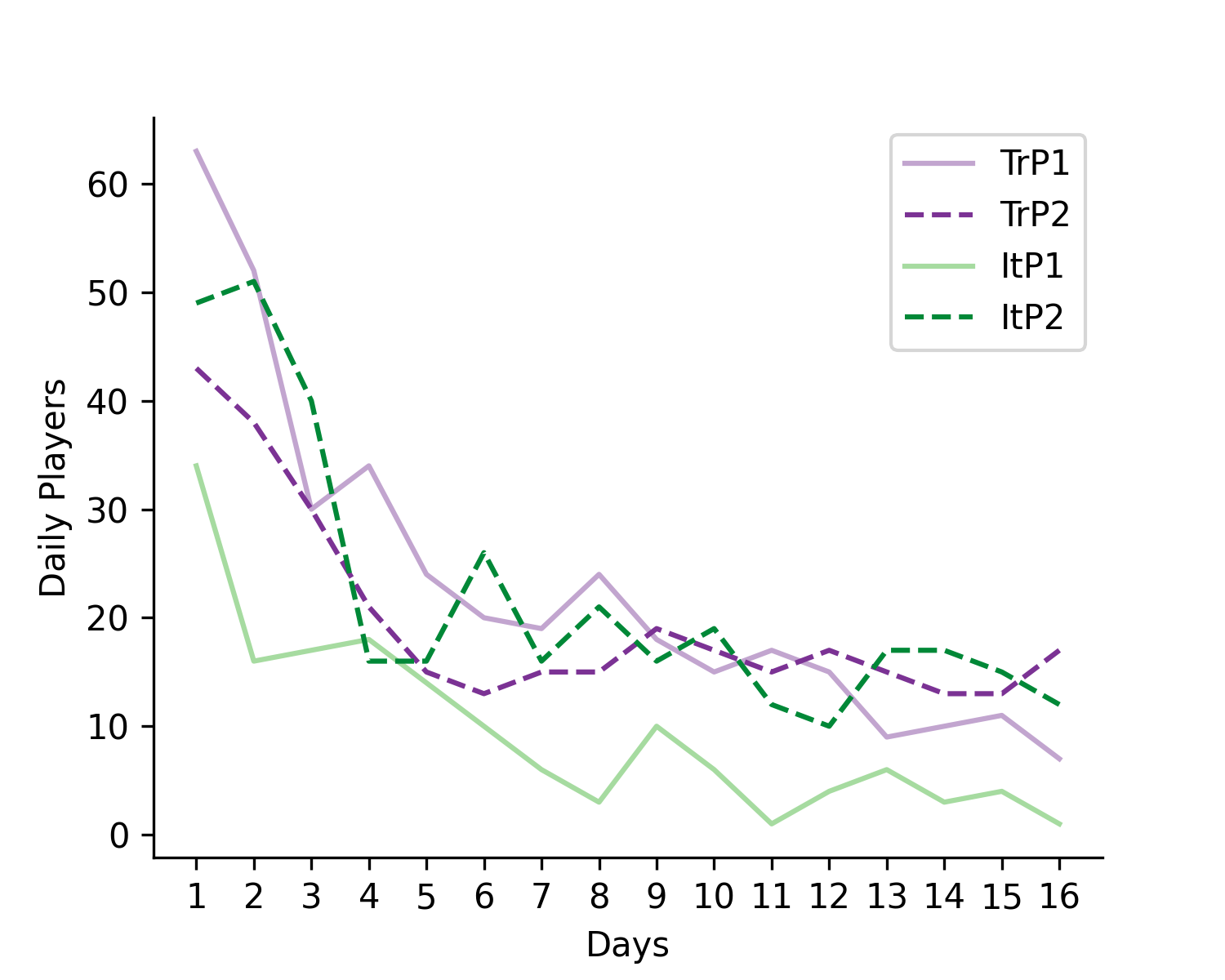}
	    \caption{Daily player count}
	    \label{fig:daily_players}
	\end{subfigure}
	\caption{Daily play statistics}
	\label{fig:daily_new_users}
\end{figure}

Figure \ref{fig:daily_new_players} shows the newly player counts for each day. 
This graphic shows the users visiting the bot, whether they start playing or not.
It can be seen that the player counts in the initial days are very high for almost all periods due to the social media announcements. 
The new player counts in the following days are relatively low compared to the initial days, which is understandable. Still, it may be seen that the game continues to spread except for ItP1.
It is worth pointing out that the spread also applies to Turkish although there had been no daily announcements contrary to Italian.

Figure \ref{fig:daily_players} provides the daily player counts who either submitted or reviewed.
It should be noted that the initial values between Figure \ref{fig:daily_new_players} and Figure \ref{fig:daily_players} differ from each other since some players, although entering the game (contributed to the new player counts in Figure \ref{fig:daily_new_players}), did not play it, or the old players from previous periods continued to play the game. 
As Figure \ref{fig:daily_players} shows, for TrP1, TrP2 and ItP2 there are more than 10 players playing the game each day (except the last day of TrP1). For ItP1, the number of daily players is under 10 for 9 days out of 16. Figure \ref{fig:daily_players} shows a general decline in daily player counts for TrP1 and ItP1, whereas each day, nearly 20 players played the game for TrP2 and ItP2.



The following constructs are selected for the analysis of the motivational and behavioral outcomes of the proposed gamification approach: 
\textit{system usage}, \textit{engagement}, \textit{loyalty}, 
\textit{ease of use}, \textit{enjoyment}, \textit{attitude}, \textit{motivation}, and \textit{willingness to recommend} \citep{morschheuser2017gamified,MORSCHHEUSER20197}. 
 These constructs are evaluated quantitatively and qualitatively via different operational means; i.e., survey results, bot usage statistics, and social media interactions.

\subsection{Data Analysis}
\label{sec:an}

\begin{figure}[!htb]
	\centering
	\begin{subfigure}[b]{0.48\textwidth}
	    \includegraphics[width=\textwidth]{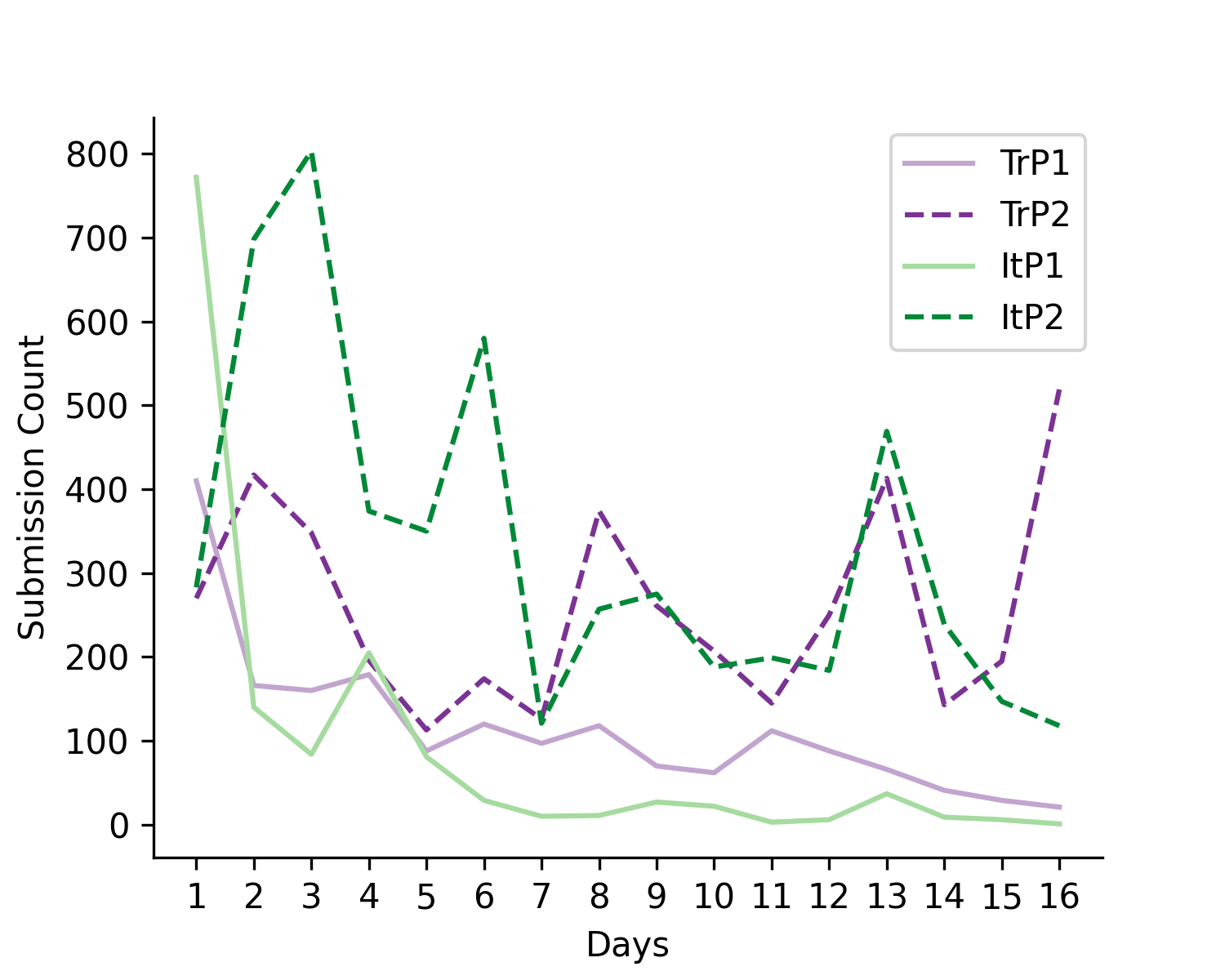}
	    \caption{Daily submission counts}
	    \label{fig:daily_submissions}
	\end{subfigure}
	\begin{subfigure}[b]{0.48\textwidth}
	    \includegraphics[width=\textwidth]{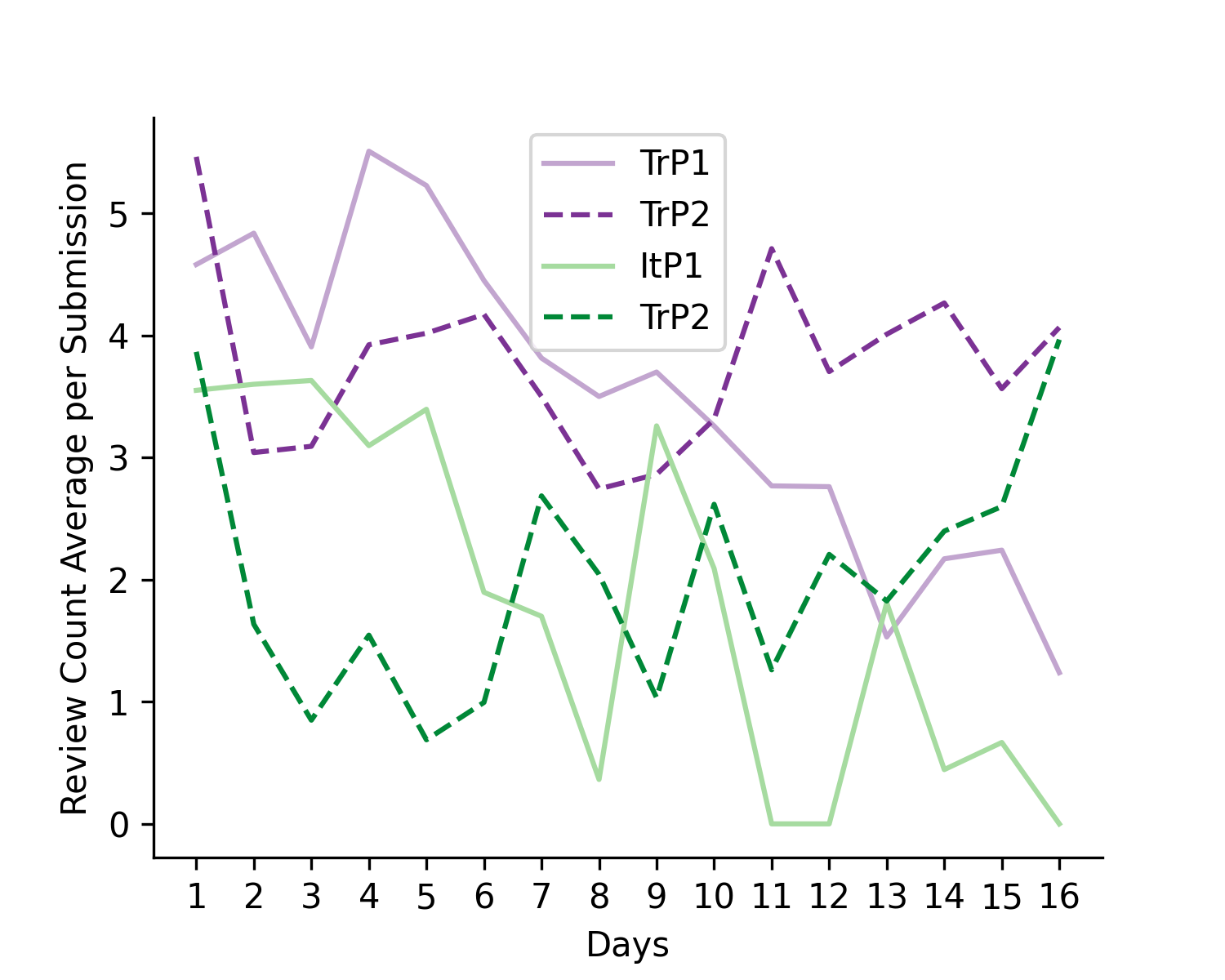}
	    \caption{Daily review count average per submission}
	    \label{fig:daily_review_average}
	\end{subfigure}
	\caption{Daily statistics for submissions and reviews}
	\label{fig:usagestatistics}
\end{figure}

\begin{figure}[!htb]
	\centering
	\begin{subfigure}[t]{0.48\textwidth}
	    \includegraphics[width=\textwidth]{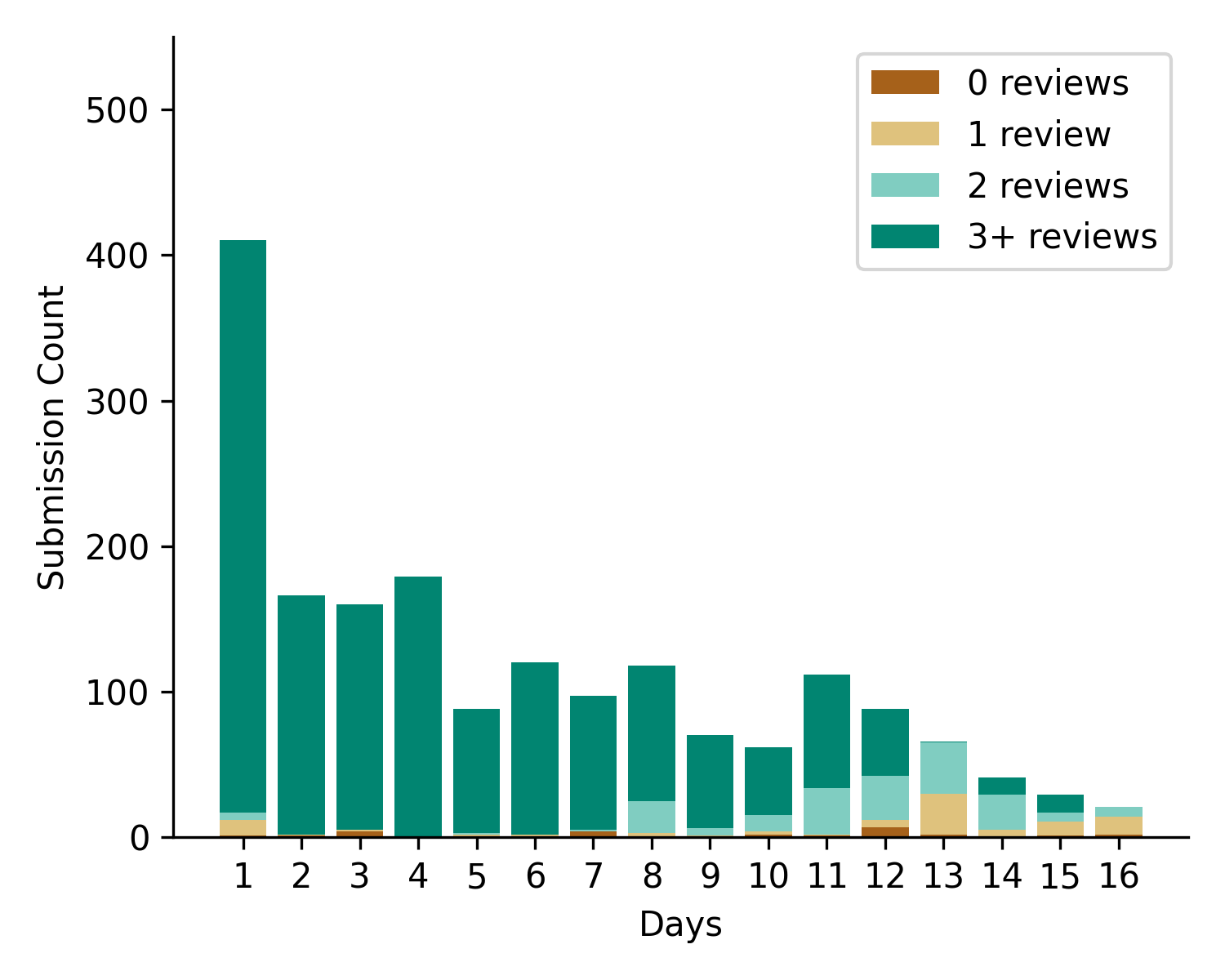}
		\caption{TrP1}
	    \label{fig:all_daily_review_count_trp1}
	\end{subfigure}
	\begin{subfigure}[t]{0.48\textwidth}
	    \includegraphics[width=\textwidth]{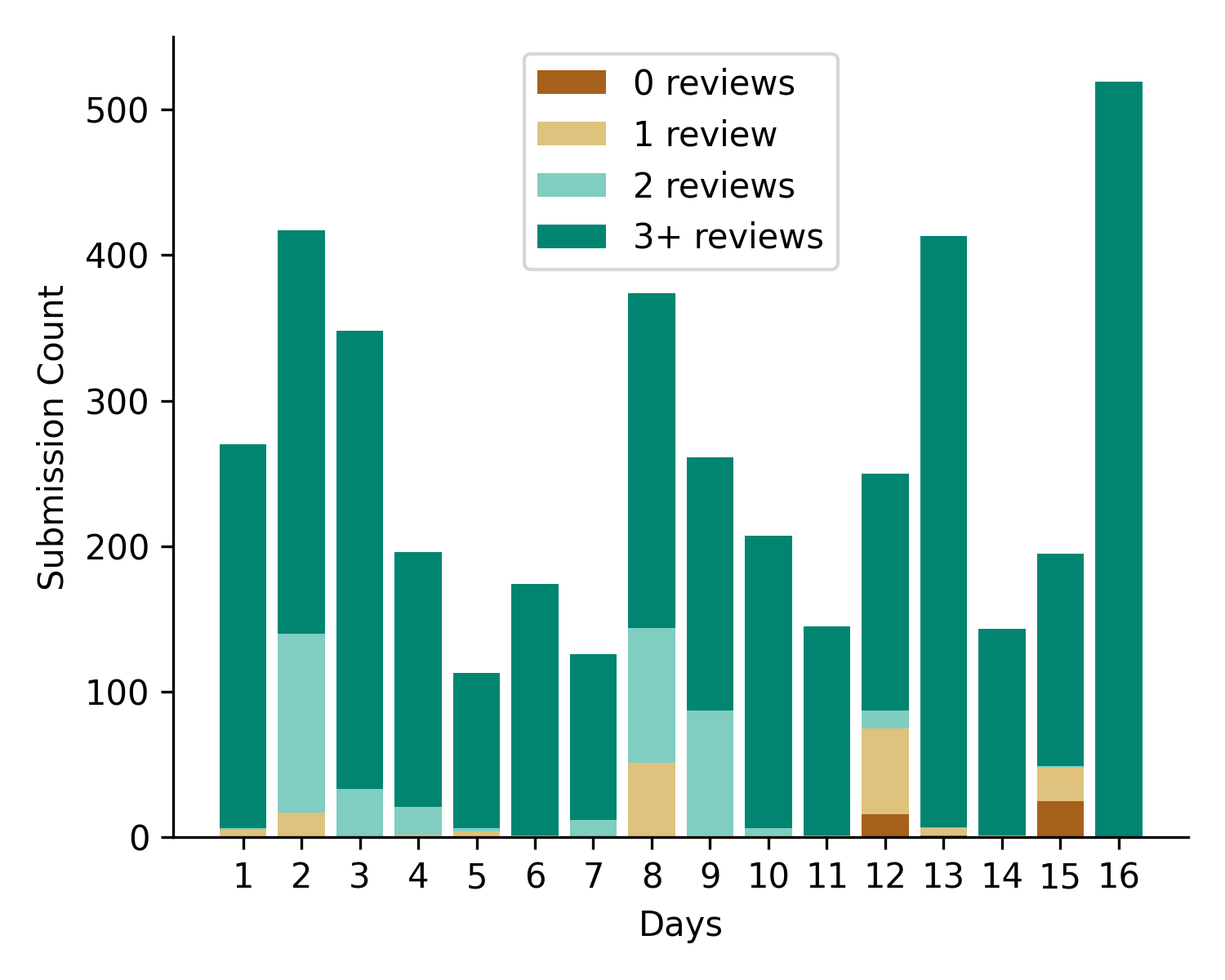}
		\caption{TrP2}
	   \label{fig:all_daily_review_count_trp2}
	\end{subfigure}
	\begin{subfigure}[b]{0.48\textwidth}
	    \includegraphics[width=\textwidth]{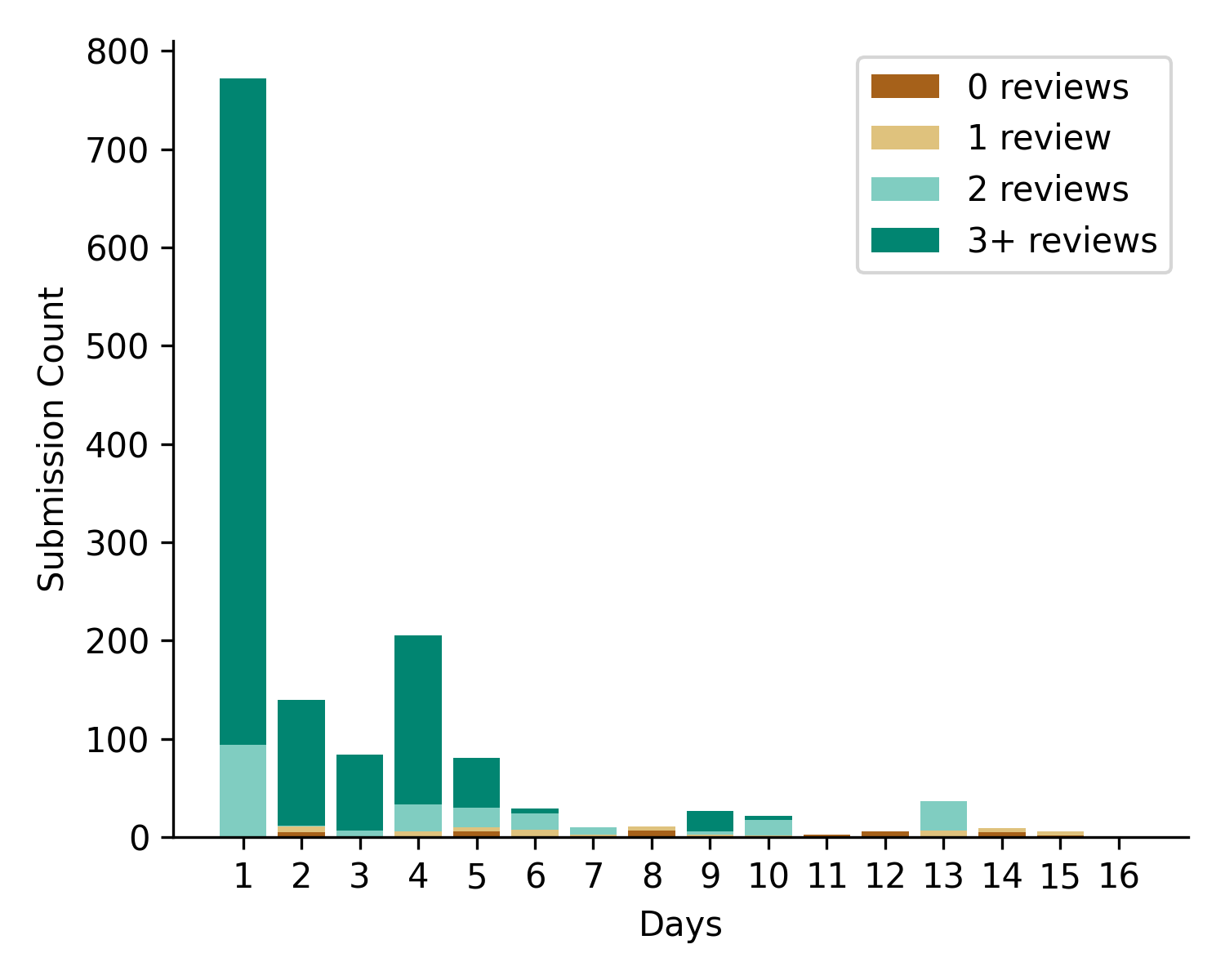}
		\caption{ItP1}
	   \label{fig:all_daily_review_count_itp1}
	\end{subfigure}
	\begin{subfigure}[b]{0.48\textwidth}
	    \includegraphics[width=\textwidth]{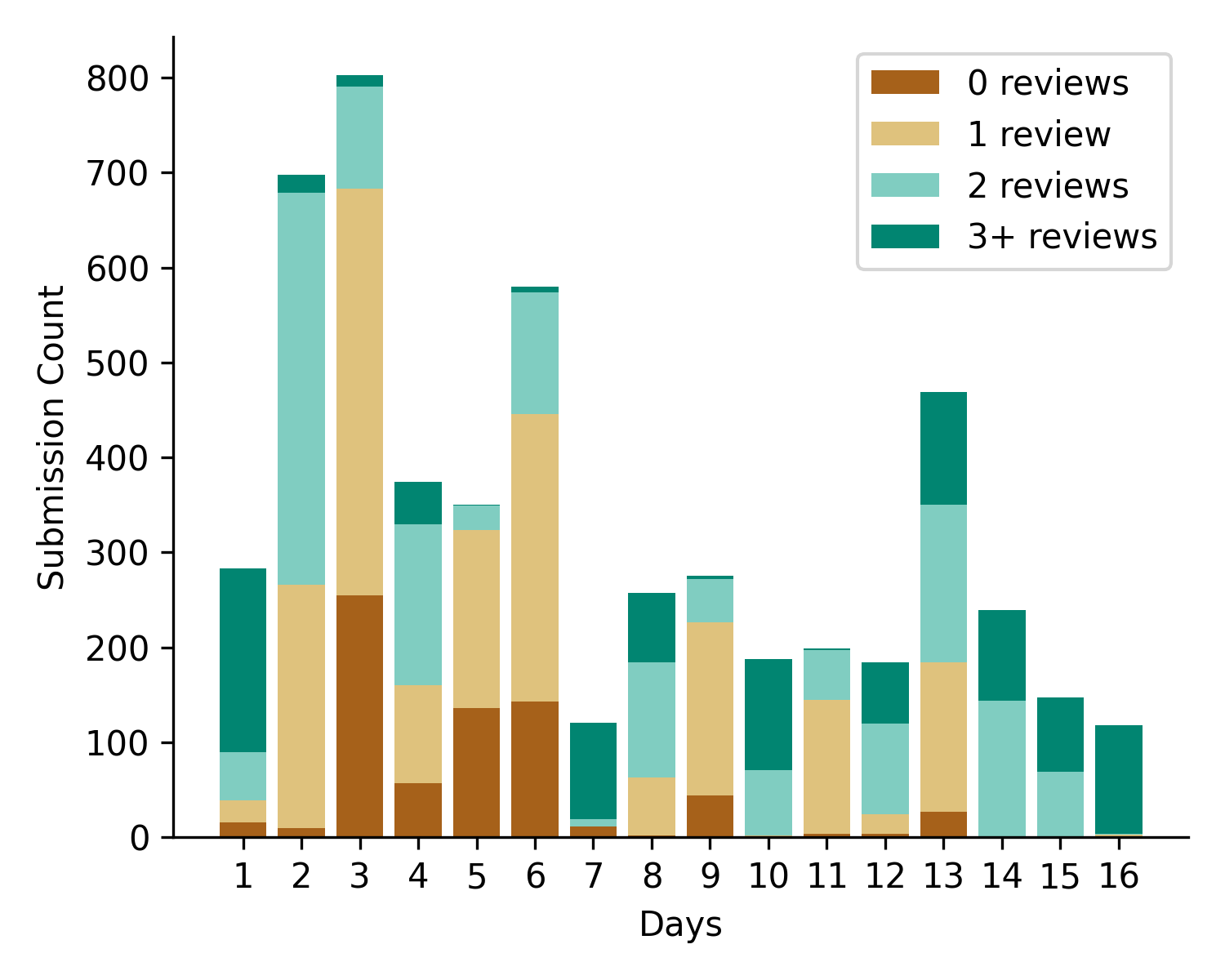}
		\caption{ItP2}
	    \label{fig:all_daily_review_count_itp2}
	\end{subfigure}
	\caption{Daily review frequencies per submission}
    \label{fig:all_daily_review_count}
\end{figure}

\begin{figure}[!htb]
	\centering
	\begin{subfigure}[t]{0.48\textwidth}
	    \includegraphics[width=\textwidth]{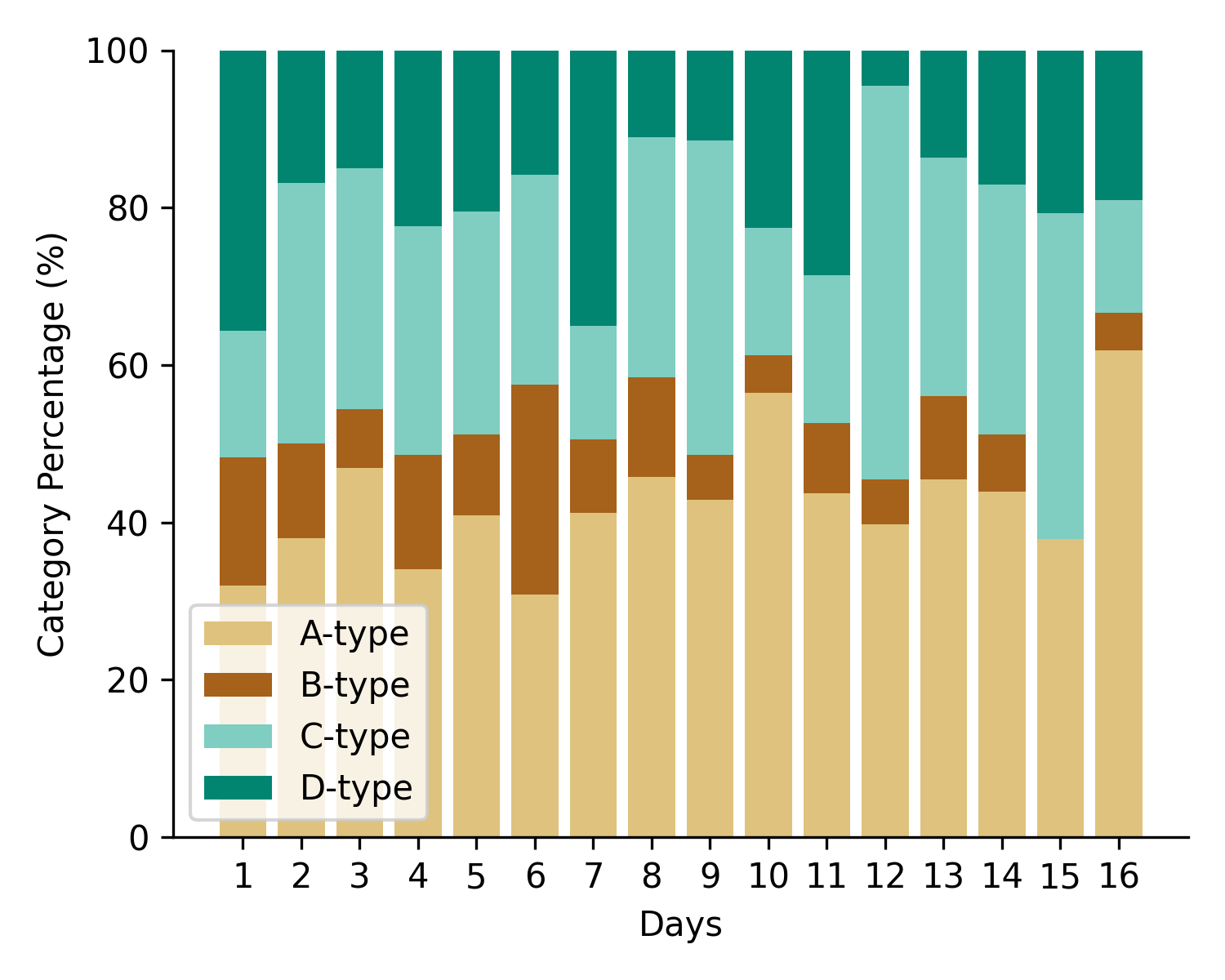}
		\caption{TrP1}
	    \label{fig:sampletypedistribution_trp1}
	\end{subfigure}
	\begin{subfigure}[t]{0.48\textwidth}
	    \includegraphics[width=\textwidth]{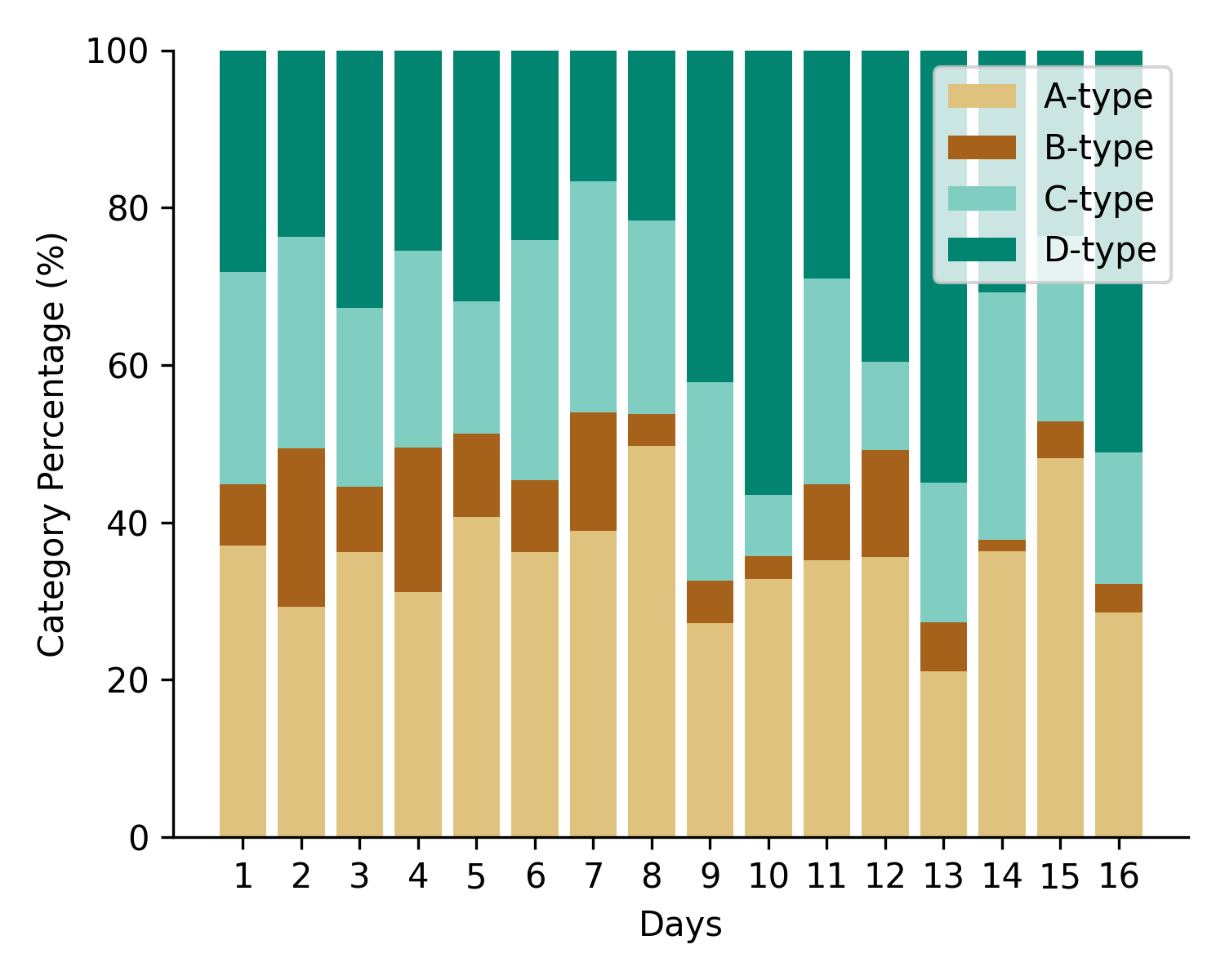}
		\caption{TrP2}
	   \label{fig:sampletypedistribution_trp2}
	\end{subfigure}
	\begin{subfigure}[b]{0.48\textwidth}
	    \includegraphics[width=\textwidth]{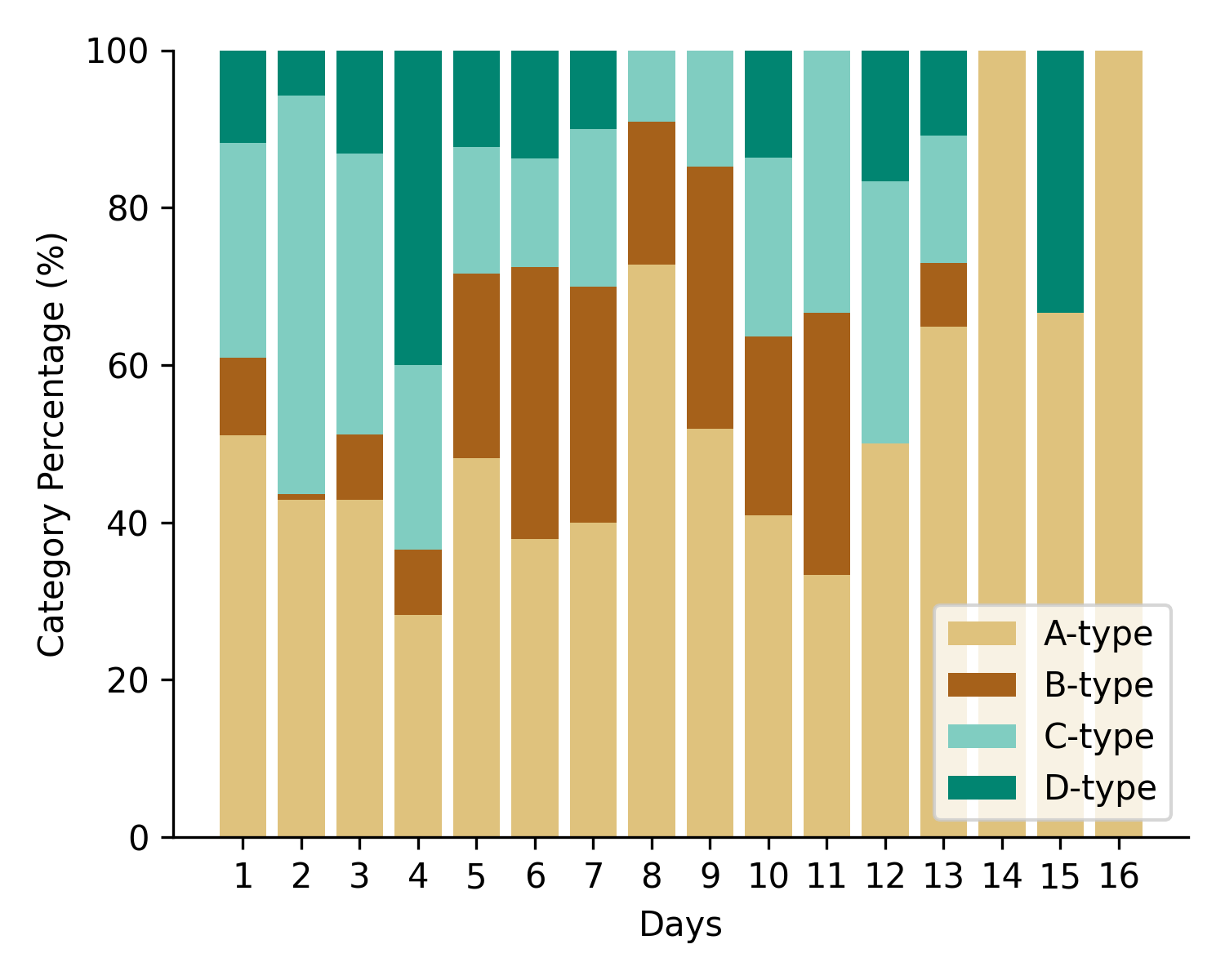}
		\caption{ItP1}
	   \label{fig:sampletypedistribution_itp1}
	\end{subfigure}
	\begin{subfigure}[b]{0.48\textwidth}
	    \includegraphics[width=\textwidth]{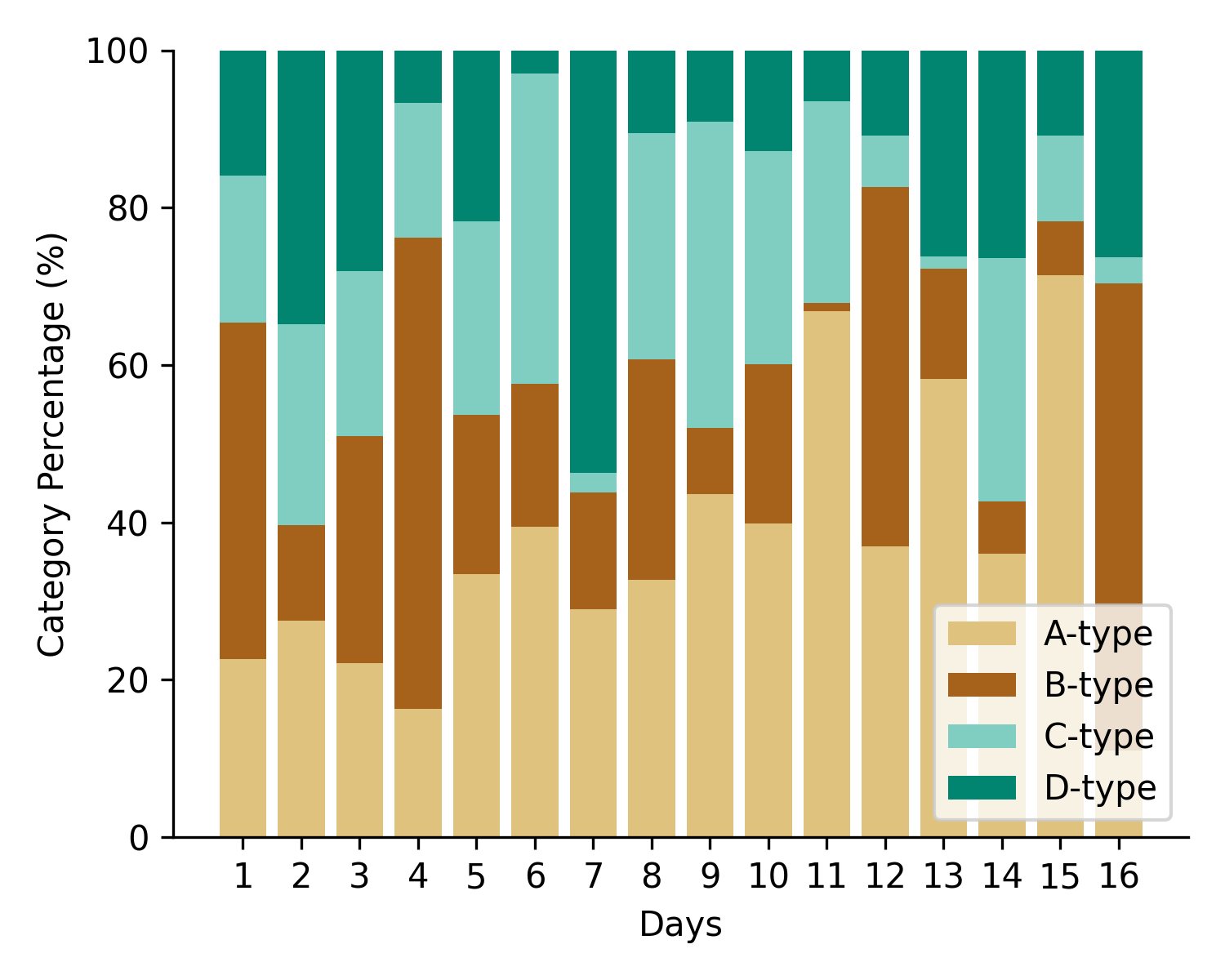}
		\caption{ItP2}
	    \label{fig:sampletypedistribution_itp2}
	\end{subfigure}
	\caption{Daily sample type distributions}
    \label{fig:sampletypedistribution}
\end{figure}

During the four 16-day periods, we collected 5978 submissions and 22712 reviews for Turkish, and 6728 submissions and 13620 reviews for Italian in total. 
In this section, we make a data analysis by providing 1) submission and average review statistics in Figure \ref{fig:usagestatistics}, 2) daily review frequencies per submission in Figure \ref{fig:all_daily_review_count}, and 3) collected sample distributions in Figure \ref{fig:sampletypedistribution} according to the sample categories provided in $\mathsection$\ref{sec:gamification}.
The impact of the monetary reward can be observed on all figures, but the comparisons between periods with and without monetary reward are left to be discussed in $\mathsection$\ref{sec:survey} under the related constructs. 
In this section, although the analyses are provided for all the four periods, the discussions are mostly carried out on TrP2 and ItP2, which yielded a more systematic data collection (see Figure~\ref{fig:daily_submissions} - Daily submission counts).

Figure \ref{fig:daily_submissions} shows that the soft target of 100 submissions per idiom is reached for both of the languages, most of the time by a large margin: 258 submissions on daily average for Turkish and 330 submissions for Italian. 
The average review counts are most of the time above 3 for Turkish idioms with a mean and its standard error of 3.7$\pm$0.2, whereas for Italian this is 2.0$\pm$0.2. The difference between averages may also be attributed to the crowd type (mostly AI-related people for Turkish, students for Italian), which is again going to be discussed in the next section. 
But in here, we may say that in ItP2, especially in the first days, the submission counts were quite high and review averages remained relatively lower when compared to this. 
However, since we have many samples on some days, although the review average is low, we still have many samples that have more than 2 reviews.
Figure \ref{fig:all_daily_review_count} shows the review count distributions per submission.
As an example, when we look at Figure \ref{fig:all_daily_review_count_itp2} the 3$^{rd}$ day of ItP2 (which received 803 samples with 0.8 reviews in average Table~\ref{tab:ITRun}), we may see that we still have more than 100 hundred samples (specified with green colors) which received more than 2 reviews. On the other hand, TrP2 results (Figure \ref{fig:all_daily_review_count_trp2}) show that there are quite a lot of submissions that are reviewed by at least 3 people.
Similarly for TrP1 (Figure \ref{fig:all_daily_review_count_trp1}) and ItP1 (Figure \ref{fig:all_daily_review_count_itp1}), although the submissions counts are lower, most of them are reviewed by at least 2 people.


The Appendix tables (Table~\ref{tab:TRrun} and Table~\ref{tab:ITRun}) also provide the dislikes percentages for each idiom in their last column.
The daily averages are 15.5$\pm$2.7 per cent for TrP2 and 24.1$\pm$3.2 per cent for ItP2.
It is worth pointing out that two days (6$^{th}$ and 11$^{th}$) in ItP2 were exceptional, and the dislike ratios were very high.
In those days there were players who entered very similar sentences with slight differences, and reviewers caught those and reported. 
It was also found that these reported players repeatedly sent dislikes to other players' entries.
The moderators had to ban them, and their submissions and reviews were excluded from the statistics. 
No such situation had been encountered in TrP2 where the idiom with the highest dislike ratio appears in the 8$^{th}$ day with 36 per cent.
Although the data aggregation stage \citep{hung2013evaluation} is out of the scope of this study, 
it is worth mentioning that despite this ratio, we still obtained many fully liked examples (87 out of 374 submissions, liked by at least 2 people).


Figure \ref{fig:sampletypedistribution} shows the type distributions (introduced in $\mathsection$\ref{sec:gamification}) of the collected samples.
We see that the used scoring, notifications, and tips helped to achieve our goal of collecting samples from various types. 
The type ratios change from idiom to idiom according to their flexibility. 
When we investigate B-type samples for Italian, we observe pronouns, nouns and mostly adverbs  intervening between the idiom components.
Italian B-type samples seem to be more prevalent than Turkish.
This is due to the fact that possessiveness is generally represented with possessive suffixes in Turkish and we don't see any B-type occurrences due to this. The possessive pronouns, if present, occur before the first component of the idiom within the sentence.
For Turkish, we see that generally interrogative markers (enclitics), adverbs, and question words intervene between the idiom components. We see that some idioms can only take some specific question words whereas others are more flexible.

As explained at the beginning, collecting samples with different morphological varieties was also one of our objectives for the aforementioned reasons.
When we investigate the samples, we observe that the crowd produced quite a lot of morphologically various samples.
For example, for the two Turkish idioms ``\#4 - karşı çıkmak'' (\textit{to oppose}) and ``\#16 defterden silmek'' (\textit{to forget someone}), we observe 65 and 57 different surface forms in 167 and 97 idiomatic samples, respectively. For these idioms, the inflections were mostly on the verbs, but still, we observe that the first constituents of the idioms were also seen under different surface forms (``karşı'' (\textit{opposite}) in 4 different surface forms inflected with different possessive suffixes and the dative case marker, and ``defterden'' (\textit{from the notebook}) in 5 different surface forms inflected with different possessive suffixes with no change on the ablative case marker). We also encounter some idioms where the first constituent only occurs under a single surface form (e.g., ``\#8 sıkı durmak'' (\textit{to stay strong or to be ready})).
The observations are in line with the initial expectations, and the data to be collected with the proposed gamification approach is undeniably valuable for building knowledge bases for idioms.


\subsection{Comparison with Parseme Annotations}
\label{sec:comp}

Parseme multilingual corpus of verbal multiword expressions \citep{Savarytv,ramisch2018edition} contains 280K sentences from 20 different languages including Italian and Turkish. As the name implies, this dataset includes many different types of verbal MWEs as well as verbal idioms. Thus, it is very appropriate to be used as an output of a classical annotation approach.
During the preparation of this corpus, datasets, retrieved mostly from newspaper text and Wikipedia articles,
were read (i.e., scanned) and annotated by human annotators according to well-defined guidelines. Since MWEs are random in these texts, only the surrounding text fragments (longer than a single sentence) around the annotated MWEs were included in the corpus, instead of the entire scanned material \citep{Savarytv}.
Due to the selected genre of the datasets, it is obvious that many idioms, especially the ones used in colloquial language, do not appear in this corpus.
Additionally, annotations are error-prone as stated in the previous sections.
Table \ref{tab:parsemelang} provides the statistics for Turkish and Italian parts of this corpus.


\begin{table}[!htp]
\centering
\caption{Parseme Turkish and Italian Datasets \citep{Savarytv}}
\label{tab:parsemelang}
\begin{tabular}{lll}
\hline
 & Turkish & Italian \\ \hline
\# of annotated sentences & 18036 & 17000 \\
\# of MWE annotations (\% of first row) & 6670 (37\%) & 2454 (14\%) \\
\# of unique MWE and their percentages\footnote{\hl{The values were calculated by counting the idioms having the same lemma information within the corpus.}?} & 1981 (32\%) & 1671 (68\%) \\
most frequent MWE count & 127 & 110 \\
VID (Verbal Idioms) & 3160 & 1163 \\
LVC (Light-Verb Constructions) & 2823 & 482 \\
VPC (Verb-Particle Constructions) & 0 & 73 \\ \hline
\end{tabular}
\end{table}

In order to compare the outputs of the classical annotation approach and the gamified construction approach, we select 4 idioms for each language (from Table~\ref{tab:TRrun} and Table~\ref{tab:ITRun}) and manually check their annotations in the Parseme corpus. 
For Turkish, we select one idiom which is annotated the most (``yer almak'' - \textit{to occur} 132 times) in the Parseme corpus, one which appears very few (``zaman öldürmek'' - \textit{to waste time} 1 time) and two which appear in between. The selected idioms are given in Table~\ref{tab:parseme}. For Italian, since the idioms were selected randomly (as opposed to Turkish ($\mathsection$\ref{sec:partic})),
their occurrence in the Parseme corpus is very rare as may be seen from the tables. Thus, we selected the 4 idioms with the highest counts. 

As stated before, only the idiomatic samples were annotated in the Parseme corpus.
To further analyze, we retrieved all the unannotated sentences containing the lemmas of the idioms' constituents and 
checked to see whether they are truly nonidiomatic usages or are mistakenly omitted by human annotators (i.e., False negatives (Fn)).
As may be seen from Table~\ref{tab:parseme}, the mistakenly omitted idiomatic samples (the last column) are quite high although this dataset is reported to be annotated by two independent research groups in two consecutive years: e.g., 16 idiomatic usage samples for 
the idiom ``meydana gelmek'' (\textit{to happen}) were mistakenly omitted out of 25 unannotated sentences. 
Similar to the findings of \citet{bontcheva2017crowdsourcing} on named entity annotations,
these results support our claim about the quality of the produced datasets when the crowd focuses on a single phenomenon at a time.
Additionally, the proposed gamified approach (with a crowdrating mechanism) also provides multiple reviews on the crowdcreated dataset.

\begin{table}[htp]
\caption{Comparison with classical data annotation. (Id.: \# of idiomatic samples, Nonid.: \# of nonidiomatic samples, Rev.: Average review count, Unnon.: \# of unannotated sentences containing lemmas of the idiom components, Fn.: \# of False Negatives, please see Table~\ref{tab:TRrun} and Table~\ref{tab:ITRun} for the meanings of idioms)}
\label{tab:parseme}
\center

\begin{tabular}{ll|ccc|ccc}
                   &           & \multicolumn{3}{c}{Dodiom}                                                       & \multicolumn{3}{|c}{Parseme}                              \\
Lang.                    & Idiom                 & Id.                     & Nonid.                  & Rev.                     & \multicolumn{1}{c}{Id.} & Unann.                 & Fn.   \\\hline
\multirow{5}{*}{Turkish} & yer almak      & 69  & 49  & 3.5 &  132   & 83 & 30     \\
                         & meydana gelmek & 103 & 92  & 3.6 &  29  & 25 & 16     \\
                         & karşı çıkmak   & 167 & 352 & 4.1 &  27  & 46 & 14     \\
                         & zaman öldürmek & 123 & 127 & 3.7 &    1 & 5 & 0     \\\hline
\multirow{5}{*}{Italian}        
& aprire gli occhi &143&132&1.0&2&0&0\\
& prendere con le pinze&409&394&0.8&1&0&0\\
& essere tra i piedi&152&32&2.2&0&3&0\\
& mandare a casa &102&137&2.4&2&2&0\\
                         \hline
\end{tabular}
\end{table}

When the idiomatic annotations in Parseme are investigated, it is seen that they are almost all A-types samples, and B-type samples very rarely appear within the corpus, which could be another side effect of the selected input text genres.

\subsection{Motivational \& Behavioral Outcomes}
\label{sec:survey}
In this section, we provide our analysis on motivational and behavioral outcomes of the proposed gamification approach for idiom corpora construction. 
The survey results (provided in Table~\ref{tab:survey}), bot usage statistics (provided in $\mathsection$\ref{sec:an}), and social media interactions are used during the evaluations. The investigated constructs are \textit{system usage}, \textit{engagement}, \textit{loyalty}, \textit{ease of use}, \textit{enjoyment}, \textit{attitude}, \textit{motivation}, and \textit{willingness to recommend}.

\begin{table}[!htp]
\caption{Survey Constructs, Questions \& Results. (Answer Types:  5 point Likert scale (5PLS),  predefined answer list (PL),  PL including the ``other'' option with free text area  (PLwO))}
\label{tab:survey}
\begin{tabular}{llp{5cm}lll}
\hline
Q & Constructs & Survey Questions & Answer Type & Turkish & Italian \\ \hline
1 & demographic & -What is your educational background? PL:\{from 1:primary school to 5:PhD\} & PL & 0 0 9 12 4 & 0 2 6 20 3 \\
2 & demographic & -What field do you work in? PLwO:\{education, AI, computer tech., other\} & PLwO & 0 12 8 5 & 6 2 2 21 \\
3 & demographic & -How old are you? PL:\{\textless{}18, 18-25, 25-30, \textgreater{}30\} & PL & 0 9 9 7 & 0 14 12 5 \\
4 & demographic & -How did you hear about Dodiom? PLwO:\{Linkedin, Twitter, a friend, other\} & PLwO & 7 4 10 4 & 6 7 9 9 \\
5 & attitude & -What's your opinion about Dodiom? & 5PLS & 0 0 0 10 15 & 0 2 1 15 12 \\
6 & motivation & -Why did you play Dodiom, what was the main motivation for you to play? PLwO:\{help dodo, daily achievements, fun,  help NLP studies, other\} & PLwO & 0 4 0 20 1 & 1 4 4 21 1 \\
7 & motivation & -The Gift Certificate was an important motivation for me to play the game. & 5PLS & 4 2 7 4 8 & 6 1 6 8 10 \\
8 & enjoyment & -The leaderboard and racing components made the game more fun. & 5PLS & 0 1 2 5 17 & 3 2 5 7 14 \\
9 & engagement & -Dodo's messages about my place in the rankings increased my participation in the game. & 5PLS & 1 0 4 9 11 & 4 3 3 8 13 \\
10 & attitude & -I liked the interface of the game and the ease of play, it kept me playing the game. & 5PLS & 0 1 0 5 19 & 0 0 9 10 12 \\
11 & ease of use & -I was able to learn the gameplay of the game without much effort. & 5PLS & 0 0 1 2 22 & 0 0 1 7 23 \\
12 & engagement & -The frequency of Dodo's notifications was not disturbing. & 5PLS & 4 2 8 3 8 & 4 4 10 7 6 \\
13 & enjoyment & -The theme and gameplay was fun, I enjoyed playing. & 5PLS & 0 0 1 8 16 & 0 1 4 11 15 \\
14 & loyalty & -Dodo will take a break from learning soon. Do you want to continue helping when it starts again? PLwO:\{yes, no, other\} & PLwO & 24 0 1 & 28 2 1 \\
15 & attitude & -Which aspect of the game did you like the most? & free-text & - & - \\
17 & attitude & -Was there anything you didn't like in the game, and if so, what? & free-text & - & - \\
18 & loyalty & -How many days did you play Dodiom? PL:\{1, 2-3, \textless{}1week, \textgreater{}1 week\} & PL & 2 2 4 16 & 7 10 7 7 \\
19 & loyalty & -How many samples did you send to Dodiom per day on average? PL:\{2-3, \textless{}10, 10-20, \textgreater{}20\} & PL & 3 7 6 9 & 12 6 7 5 \\
20 & - & -Can you share any suggestions about the game? & free-text & - & - \\ \hline
\end{tabular}
\end{table}
\pagestyle{plain}

Table~\ref{tab:survey} summarizes the survey results in terms of response counts provided in the last two columns for Turkish and Italian games, respectively. In questions with 5 point Likert scale answers, the options go from 1: strongly disagree or disliked to 5: strongly agree or liked.
The first 4 questions of the survey are related to demographic information. The answers to question 2 (Q2 of Table~\ref{tab:survey}) reveal that the respondents for Turkish play are mostly AI and computer technology related people (21 out of 25 participants selected the related options and 2 stated NLP under the \textit{other} option), whereas for Italian play they are from different backgrounds; 21 people out of 31 selected the \textit{other} option where only 2 of them stated NLP and computational linguistics, and the others gave answers like translation, student, administration, tourism, and sales.
The difference between crowd types seems to also affect their behavior. In TrP2, we observe that the review ratios are higher than ItP2 as stated in the previous section. On the other hand, ItP2 participants made more submissions. There were more young people in Italian plays (Q3) than Turkish plays. The appearing situation may be attributed to their eagerness to earn more points. We had many free text comments (to be discussed below) related to the low scoring of the review process from both communities.

The overall \textit{system usage} of the participants is provided in $\mathsection$\ref{sec:an}.
Figure \ref{fig:daily_players} and Figure \ref{fig:usagestatistics} shows player counts and their play rates.
Although, only 50 per cent of survey Q7 answers, about the gift card \textit{motivation}, says agree (4) or strongly agree (5),
the graphics mentioned above reveal that the periods with additional incentives (i.e., gift card rewards) (TrP2, ItP2) are more successful at fulfilling the expectations about \textit{loyalty} than the periods without (TrP1, ItP1).
Again related to the  \textit{loyalty} construct (Q18 and Q19), we see that more than half of the Turkish survey participants were playing the game for more than 1 week at the time of filling out the survey (which was open for the last three days of TrP2) and they were providing more than 10 samples each day. Since the Italian survey was open for a longer period of time (see Table \ref{tab:user}), we see a more diverse distribution on the answers.
Most of the participants also stated that they would like to continue playing the game (Q14).

A very high number of participants (20 out of 25, and 21 out of 31) stated that their \textit{motivation} to play the game was to help NLP studies. 4 of them answered Q15 as: \textit{``-I felt that I'm helping a study.'', ``-The scientific goal'', ``-The ultimate aim'', ``-I liked it being the first application designed to provide input for Turkish NLP as far as I know. Apart from that, we are forcing our brains while 
entering in a sweet competition with the friends working in the field and contributing at the same time.''}
We see that the gamification elements and the additional incentive helped the players to stay on the game with this motivation (Q8, Q13 \textit{enjoyment}). 
In TrP2, we also observed that some game-winners shared their achievements from social media (\textit{willingness to recommend}) and found each other from the same channel. 
Setting more moral goals than monetary rewards, they combined distributed bookstore gift cards and sent book gifts to poor children by using these. Around 800 social media likes and shares were made in total (for both languages).
More than half of the respondents chose the answer ``from a friend'' or ``other'' to Q4 (``How did you hear about Dodiom?'') instead of the first two options covering Linkedin and Twitter. The ``other'' answers were covering either the name of the influencers, or Facebook and Instagram for Italian. We may say that the spread of the game (introduced in $\mathsection$\ref{sec:partic}) is not due to the social media influences alone but people let each other now about it, which could also be seen as an impact of their \textit{willingness to recommend} the game.

Almost all of the users found the game easy to understand and play (Q11 \textit{ease of use}). 
Almost all of them liked the game; only 3 out of 31 Italian participants scored under 4 (liked) to Q5 (\textit{attitude}), and 9 of them scored neutral (3) to Q10. Only 1 Turkish participant was negative to this later question about the interface.
When we analyze  the free-text answers to Q15, we see that 8 people out of 56 stated that they liked the game elements. Some answers are as follows: \textit{``-I loved Dodo gifts.'', ``-Gamification and story was good'', \mbox{``-it} triggers a sense of competition'', ``-The icon of the application is very sympathetic, the messages are remarkable.'', ``-I liked the competition content and the ranking'', ``-gift voucher'', ``-interaction with other players''}.
Three participants stated that they liked the game being a bot with no need to download an application.
Three of them mentioned that they liked playing the game: \textit{``-I liked that the game increases my creativeness. It makes me think. I'm having fun myself.'', ``-To see original sentences'', ``-... Besides these, a fun opportunity for mental gymnastics.'', `-`Learn new idioms'', ``-Linguistic aspect''}.
8 participants stated that they liked the uniqueness of the idea: \textit{``-The originality of the idea'', ``-the creativity'', ``-Efficiency and immediacy'', ``-The chosen procedure'', ``-the idea is very useful for increasing the resources for the identification of idiomatic expressions.'', ``-The idea of being interacting with someone else'', ``-Undoubtedly, as a Ph.D. student in the field of NLP, I liked that it reduces the difficulty of labeling data, makes it fun, and is capable of enabling other people to contribute whether they are from the field or not''}.

More than half of the participants were okay with the frequency of the Dodo's instant messages and most of them agreed about their usefulness in keeping them in the game (Q9 and Q12). 4 people out of 56 participants in total complained about the frequency of the messages as an answer to Q16 (\textit{``-Slightly frequent notifications'', ``-Notifications can be sent less often.'',``-Too many notifications''}).
As opposed to this, one participant said \textit{``It's nice that when you put it aside, the reminders and notifications that encourage you to earn more points make me re-enter words during the day''} as an answer to Q15.

Other answers to Q16 are as follows: \textit{``-I don't think it should allow the possibility of repeating the same sentences.'', ``-It can get repetitive, a mixed-mode where automatically alternating between suggestions and evaluations with multiple expressions per day would have been more engaging'', ``-Error occurrence during voting has increased recently. Maybe it could be related to increased participation. However, there is no very critical issue.'', ``-Sometimes it froze''}. Regarding the last two comments, we have stated in the previous sections the need for optimization towards the end of the play with the increased workload and the action taken. On the other hand, the first two comments are also very good indicators for future directions.

For Q19, we received 3 suggestions for the scoring system, 1 suggestion for automatic spelling correction, 2 suggestions for detailing dislikes, and 1 suggestion for the need to cancel/change an erroneous submission or review.
Obviously, the users wanted to warn about spelling mistakes in the input sentences but hesitated to send a dislike due to this. That is why they suggested  differentiating dislikes according to their reasons.
Suggestions for scoring are as follows: \textit{``-More points can be given to the reviews'', ``- The low score for reviews causes the reviewing to lose importance, especially for those who play for leadership. Because while voting, you both get fewer points and in a sense, you make your opponents earn points.'', ``-I would suggest that the score was assigned differently, that is, that the 10/15 points can be obtained when sending a suggestion (and not when others evaluate it positively). In this way, those who evaluate will have more incentives to positively evaluate the suggestions of others (without the fear of giving more points to others) (thus giving a point to those who evaluate and one to those who have been evaluated)''}. We see that in the last two comments, the players are afraid of making other players earn points. 

As explained in the game design section above, the reviews worth 1 point and sometimes 2 in happy hours, triggered by the moderators to attract the attention of the players.
Although open for discussions and changes in future trials, in the original design, we didn't want to give high points to reviews since we believe that people should review with the responsibility of a cooperative effort to create a public data set. 
Giving very high scores to reviews may lead to unexpected results.
Other scenarios together with cheating mechanisms (such as consecutive rapid likes/dislikes detection) may be considered in future works.
As stated before, we had some reporting and banning mechanisms added to control cheating/gaming the system in line with DP\#10.
The literature recommends that this is necessary since it can reverse the effects of gamification and discourage users. ``However, some experts reported that cheating could also help to better understand the users and to optimize gamification designs accordingly'' \citep{MORSCHHEUSER2018219}. As future work, automatic cheating detection for detecting rephrases and malicious reviews may be studied.

\begin{figure}[!htb]
    \centering
    \includegraphics[width=0.6\textwidth]{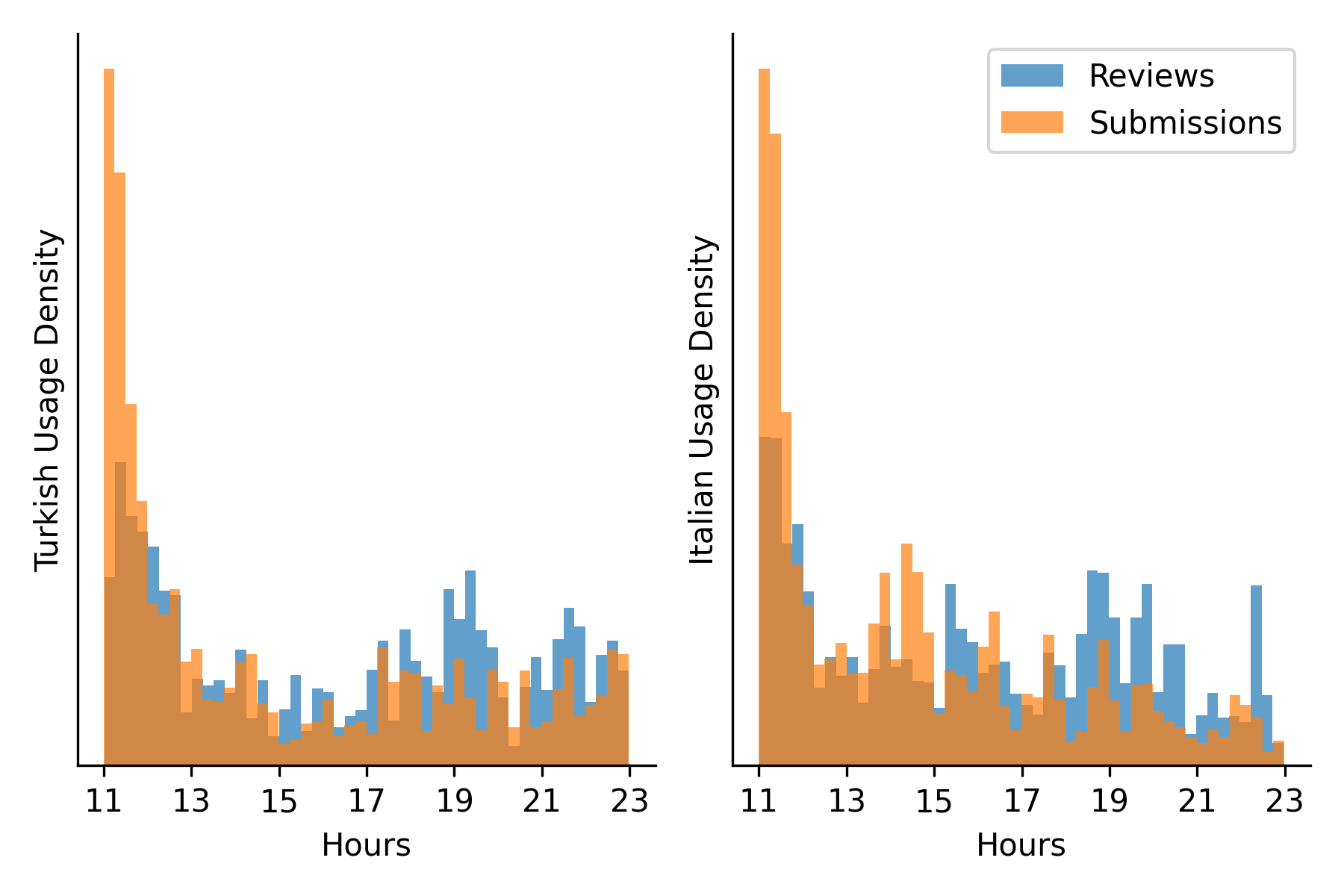}
    \caption{Histogram of interaction times in TrP2 and ItP2}
    \label{fig:daily_category_percentages_it}
\end{figure}

``Tailoring the game elements according to the users' profile is a way to improve their experience while interacting
with a gamified system, and has been noted as a current trend in gamification research'' \citep{KLOCK2020102495}.
We tested the game in an asynchronous multiplayer game scenario where the players are free to choose the time they want to contribute according to their  schedule.
Figure~\ref{fig:daily_category_percentages_it} shows the interaction times of the users, where the submissions are high at the beginning of the day and the reviews surpass the submissions towards the end of the day.
Also, the individual peaks and increased density near the end of the days correspond to the happy hour notifications sent by moderators (generally around 5 p.m. Istanbul Time for both languages, and observed as peaks around 7 p.m.  in Figure~\ref{fig:daily_category_percentages_it} Italian graphic on the right). 
However, other more condensed timings may also be considered depending on the crowd in focus. 
The game currently targets adult native speakers. 
During the initial sharing of the first prototype with the stakeholders, the initial reaction of language teachers was to use the game within classrooms as a teaching aid as well. 
For future direction, one may consider developing a different mode of the game for within classroom settings for both native speaker students or foreign language learners. The game may be enhanced to be played under the moderation of the teachers.

\section{Conclusion}Idiom corpora are valuable resources for foreign language learning, natural language  processing, and  lexicographic studies.
Unfortunately, they are rare and hard to construct. 
For the first time in the literature, this article introduced a gamified approach that uses crowdcreating and crowdrating techniques
to speed up idiom corpora construction for different languages. 
The approach has been evaluated under different motivational strategies on two languages, which produced the first idiom corpora for Turkish and Italian.
The implementation developed as a Telegram messaging bot and the collected data for the two languages in a time span of 30 days 
are shared with the researchers.
Our detailed qualitative and quantitative analyses revealed that the outcomes of the research 
are appreciated by the crowd, found useful and enjoyable,
and yielded to the collection and assessment of valuable samples that illustrate the different ways of use, which is not 
easily achievable with traditional data annotation techniques.  
Gift cards were found to be very effective in incentivizing  the users to continue playing the game in addition to gamification affordances. 


Our first short-term goal is to extend and play the game for languages other than the ones in this article, especially for languages with few  lexical resources. 	We hope that the game introduced as an opensource project will speed up the development of idiom corpora and  the research in the field.	


	
\section*{Acknowledgments}

The authors would like to offer their special thanks to Cihat Eryiğit for the discussions during the initial design of the game, 
Fatih Bektaş, Branislava Šandrih, Josip Mihaljević, Martin Benjamin, Daler Rahimjonov, and Doruk Eryiğit for fruitful discussions during its implementation, Federico Sangati for providing the codes of a telegram bot example (Plagio which is used in another game~\citep{Substituto} for foreign language learners practicing phrasal verbs), Martin Benjamin for helping on the English localization messages, 
Inge Maria Cipriano for discussion concerning the game interactions and possible deceptive behaviours by players,
and all the anonymous volunteer players. 
The study has been proposed as a task by the first author and took place in a crowdfest event (in February 2020 in Coimbra, Portugal) of EU COST Action (CA16105) Enetcollect, where the prototype has been introduced and discussed with stakeholders. The authors would like to thank Enetcollect for this opportunity which initiated new collaborations and ideas.

\bibliographystyle{nlelike}

\bibliography{res}

\begin{thebibliography}{}

\bibitem[Akkaya et~al., 2010]{akkaya2010amazon}
{\bf Akkaya, C.}, {\bf Conrad, A.}, {\bf Wiebe, J.}, \textbf{and} {\bf
  Mihalcea, R.} 2010.
\newblock Amazon mechanical turk for subjectivity word sense disambiguation.
\newblock In {\em Proceedings of the NAACL HLT 2010 workshop on creating speech
  and language data with Amazon’s Mechanical Turk}, pp. 195--203.

\bibitem[Araneta et~al., 2020]{Substituto}
{\bf Araneta, M.~G.}, {\bf Eryiğit, G.}, {\bf König, A.}, {\bf Lee, J.-U.},
  {\bf Luís, A.}, {\bf Lyding, V.}, {\bf Nicolas, L.}, {\bf Rodosthenous, C.},
  \textbf{and} {\bf Sangati, F.} 2020.
\newblock Substituto - a synchronous educational language game for simultaneous
  teaching and crowdsourcing.
\newblock In {\em Proceedings of the 9th NLP4CALL}, ~1.

\bibitem[Artignan et~al., 2009]{artignan2009multiscale}
{\bf Artignan, G.}, {\bf Hasco{\"e}t, M.}, \textbf{and} {\bf Lafourcade, M.}
  2009.
\newblock Multiscale visual analysis of lexical networks.
\newblock In {\em 2009 13th International Conference Information
  Visualisation}, pp. 685--690. IEEE.

\bibitem[Birke and Sarkar, 2006]{birke2006clustering}
{\bf Birke, J.} \textbf{and} {\bf Sarkar, A.} 2006.
\newblock A clustering approach for nearly unsupervised recognition of
  nonliteral language.
\newblock In {\em 11th Conference of the European Chapter of the Association
  for Computational Linguistics}, ~1.

\bibitem[Bontcheva et~al., 2017]{bontcheva2017crowdsourcing}
{\bf Bontcheva, K.}, {\bf Derczynski, L.}, \textbf{and} {\bf Roberts, I.} 2017.
\newblock Crowdsourcing named entity recognition and entity linking corpora.
\newblock In {\em Handbook of Linguistic Annotation}, pp. 875--892. Springer.

\bibitem[Caruso et~al., 2019]{caruso2019can}
{\bf Caruso, V.}, {\bf Barbara, B.}, {\bf Monti, J.}, \textbf{and} {\bf
  Roberta, P.} 2019.
\newblock How can app design improve lexicographic outcomes? examples from an
  italian idiom dictionary.
\newblock In {\em ELEX 2019: SMART LEXICOGRAPHY}, pp. 374--396. Lexical
  Computing CZ sro,.

\bibitem[Chamberlain et~al., 2008]{chamberlain2008addressing}
{\bf Chamberlain, J.}, {\bf Poesio, M.}, \textbf{and} {\bf Kruschwitz, U.}
  2008.
\newblock Addressing the resource bottleneck to create large-scale annotated
  texts.
\newblock In {\em Semantics in Text Processing. STEP 2008 Conference
  Proceedings}, pp. 375--380.

\bibitem[Chklovski, 2005]{chklovski2005collecting}
{\bf Chklovski, T.} 2005.
\newblock Collecting paraphrase corpora from volunteer contributors.
\newblock In {\em Proceedings of the 3rd international conference on Knowledge
  capture}, pp. 115--120.

\bibitem[Constant et~al., 2017]{mweprocessingsurvey}
{\bf Constant, M.}, {\bf Eryiğit, G.}, {\bf Monti, J.}, {\bf van~der Plas,
  L.}, {\bf Ramisch, C.}, {\bf Rosner, M.}, \textbf{and} {\bf Todirascu, A.}
  2017.
\newblock Multiword expression processing: A survey.
\newblock {\em Computational Linguistics}, 43(4):837--892.

\bibitem[Cook et~al., 2008]{cook2008vnc}
{\bf Cook, P.}, {\bf Fazly, A.}, \textbf{and} {\bf Stevenson, S.} 2008.
\newblock The vnc-tokens dataset.
\newblock In {\em Proceedings of the LREC Workshop Towards a Shared Task for
  Multiword Expressions (MWE 2008)}, pp. 19--22.

\bibitem[Dumitrache et~al., 2013]{dumitrache2013dr}
{\bf Dumitrache, A.}, {\bf Aroyo, L.}, {\bf Welty, C.}, {\bf Sips, R.-J.},
  \textbf{and} {\bf Levas, A.} 2013.
\newblock Dr. detective": combining gamification techniques and crowdsourcing
  to create a gold standard in medical text.
\newblock In {\em Proceedings of the 1st International Conference on
  Crowdsourcing the Semantic Web}, volume 1030, ~1.

\bibitem[Fort et~al., 2011]{fort2011amazon}
{\bf Fort, K.}, {\bf Adda, G.}, \textbf{and} {\bf Cohen, K.~B.} 2011.
\newblock Amazon mechanical turk: Gold mine or coal mine?
\newblock {\em Computational Linguistics}, 37(2):413--420.

\bibitem[Fort et~al., 2018]{fort2018fingers}
{\bf Fort, K.}, {\bf Guillaume, B.}, {\bf Constant, M.}, {\bf Lefebvre, N.},
  \textbf{and} {\bf Pilatte, Y.-A.} 2018.
\newblock “fingers in the nose”: Evaluating speakers’ identification of
  multi-word expressions using a slightly gamified crowdsourcing platform.
\newblock In {\em Proceedings of the Joint Workshop on Linguistic Annotation,
  Multiword Expressions and Constructions (LAW-MWE-CxG-2018)}, pp. 207--213.

\bibitem[Fort et~al., 2020]{fort-etal-2020-rigor}
{\bf Fort, K.}, {\bf Guillaume, B.}, {\bf Pilatte, Y.-A.}, {\bf Constant, M.},
  \textbf{and} {\bf Lef{\`e}bvre, N.} 2020.
\newblock Rigor mortis: Annotating {MWE}s with a gamified platform.
\newblock In {\em Proceedings of the 12th Language Resources and Evaluation
  Conference ({LREC} 2020)}, pp. 4395--4401, Marseille, France. European
  Language Resources Association.

\bibitem[Geiger and Schader, 2014]{GEIGER20143}
{\bf Geiger, D.} \textbf{and} {\bf Schader, M.} 2014.
\newblock Personalized task recommendation in crowdsourcing information systems
  — current state of the art.
\newblock {\em Decision Support Systems}, 65:3 -- 16.
\newblock Crowdsourcing and Social Networks Analysis.

\bibitem[Hashimoto and Kawahara, 2009]{hashimoto2009compilation}
{\bf Hashimoto, C.} \textbf{and} {\bf Kawahara, D.} 2009.
\newblock Compilation of an idiom example database for supervised idiom
  identification.
\newblock {\em Language resources and evaluation}, 43(4):355.

\bibitem[Howe, 2006]{howe2006rise}
{\bf Howe, J.} 2006.
\newblock The rise of crowdsourcing.
\newblock {\em Wired magazine}, 14(6):1--4.

\bibitem[Hung et~al., 2013]{hung2013evaluation}
{\bf Hung, N. Q.~V.}, {\bf Tam, N.~T.}, {\bf Tran, L.~N.}, \textbf{and} {\bf
  Aberer, K.} 2013.
\newblock An evaluation of aggregation techniques in crowdsourcing.
\newblock In {\em International Conference on Web Information Systems
  Engineering}, pp. 1--15. Springer.

\bibitem[Kaschak and Saffran, 2006]{kaschak2006idiomatic}
{\bf Kaschak, M.~P.} \textbf{and} {\bf Saffran, J.~R.} 2006.
\newblock Idiomatic syntactic constructions and language learning.
\newblock {\em Cognitive Science}, 30(1):43--63.

\bibitem[Kato et~al., 2018]{kato2018construction}
{\bf Kato, A.}, {\bf Shindo, H.}, \textbf{and} {\bf Matsumoto, Y.} 2018.
\newblock Construction of large-scale {E}nglish verbal multiword expression
  annotated corpus.
\newblock In {\em Proceedings of the Eleventh International Conference on
  Language Resources and Evaluation ({LREC} 2018)}, Miyazaki, Japan. European
  Language Resources Association (ELRA).

\bibitem[Klock et~al., 2020]{KLOCK2020102495}
{\bf Klock, A. C.~T.}, {\bf Gasparini, I.}, {\bf Pimenta, M.~S.}, \textbf{and}
  {\bf Hamari, J.} 2020.
\newblock Tailored gamification: A review of literature.
\newblock {\em International Journal of Human-Computer Studies}, 144:102495.

\bibitem[Konopka and Bock, 2009]{konopka2009lexical}
{\bf Konopka, A.~E.} \textbf{and} {\bf Bock, K.} 2009.
\newblock Lexical or syntactic control of sentence formulation? structural
  generalizations from idiom production.
\newblock {\em Cognitive Psychology}, 58(1):68--101.

\bibitem[Lawson et~al., 2010]{lawson2010annotating}
{\bf Lawson, N.}, {\bf Eustice, K.}, {\bf Perkowitz, M.}, \textbf{and} {\bf
  Yetisgen-Yildiz, M.} 2010.
\newblock Annotating large email datasets for named entity recognition with
  mechanical turk.
\newblock In {\em Proceedings of the NAACL HLT 2010 workshop on creating speech
  and language data with Amazon’s Mechanical Turk}, pp. 71--79.

\bibitem[Loper and Bird, 2002]{loper2002nltk}
{\bf Loper, E.} \textbf{and} {\bf Bird, S.} 2002.
\newblock Nltk: the natural language toolkit.
\newblock {\em arXiv preprint cs/0205028}.

\bibitem[Losnegaard et~al., 2016]{losnegaard2016parseme}
{\bf Losnegaard, G.~S.}, {\bf Sangati, F.}, {\bf Escart{\'\i}n, C.~P.}, {\bf
  Savary, A.}, {\bf Bargmann, S.}, \textbf{and} {\bf Monti, J.} 2016.
\newblock {PARSEME} survey on {MWE} resources.
\newblock In {\em Proceedings of the Tenth International Conference on Language
  Resources and Evaluation ({LREC}'16)}, pp. 2299--2306, Portoro{\v{z}},
  Slovenia. European Language Resources Association (ELRA).

\bibitem[Mitrovi{\'c}, 2013]{mitrovic2013crowdsourcing}
{\bf Mitrovi{\'c}, J.} 2013.
\newblock Crowdsourcing and its application.
\newblock {\em INFOtheca-Journal of Informatics \& Librarianship}, 14(1).

\bibitem[Morschheuser and Hamari, 2019]{morschheuser2019gamification}
{\bf Morschheuser, B.} \textbf{and} {\bf Hamari, J.} 2019.
\newblock The gamification of work: Lessons from crowdsourcing.
\newblock {\em Journal of Management Inquiry}, 28(2):145--148.

\bibitem[Morschheuser et~al., 2017]{morschheuser2017gamified}
{\bf Morschheuser, B.}, {\bf Hamari, J.}, {\bf Koivisto, J.}, \textbf{and} {\bf
  Maedche, A.} 2017.
\newblock Gamified crowdsourcing: Conceptualization, literature review, and
  future agenda.
\newblock {\em International Journal of Human-Computer Studies}, 106:26--43.

\bibitem[Morschheuser et~al., 2019]{MORSCHHEUSER20197}
{\bf Morschheuser, B.}, {\bf Hamari, J.}, \textbf{and} {\bf Maedche, A.} 2019.
\newblock Cooperation or competition – when do people contribute more? a
  field experiment on gamification of crowdsourcing.
\newblock {\em International Journal of Human-Computer Studies}, 127:7 -- 24.
\newblock Strengthening gamification studies: critical challenges and new
  opportunities.

\bibitem[Morschheuser et~al., 2018]{MORSCHHEUSER2018219}
{\bf Morschheuser, B.}, {\bf Hassan, L.}, {\bf Werder, K.}, \textbf{and} {\bf
  Hamari, J.} 2018.
\newblock How to design gamification? a method for engineering gamified
  software.
\newblock {\em Information and Software Technology}, 95:219 -- 237.

\bibitem[Murillo-Zamorano et~al., 2020]{MURILLOZAMORANO2020100645}
{\bf Murillo-Zamorano, L.~R.}, {\bf {Ángel López Sánchez}, J.}, \textbf{and}
  {\bf {Bueno Muñoz}, C.} 2020.
\newblock Gamified crowdsourcing in higher education: A theoretical framework
  and a case study.
\newblock {\em Thinking Skills and Creativity}, 36:100645.

\bibitem[{Palmero Aprosio} and {Moretti}, 2016]{palmeromorettitint}
{\bf {Palmero Aprosio}, A.} \textbf{and} {\bf {Moretti}, G.} 2016.
\newblock {Italy goes to Stanford: a collection of CoreNLP modules for
  Italian}.
\newblock {\em ArXiv e-prints}.

\bibitem[Prpić et~al., 2015]{PRPIC201577}
{\bf Prpić, J.}, {\bf Shukla, P.~P.}, {\bf Kietzmann, J.~H.}, \textbf{and}
  {\bf McCarthy, I.~P.} 2015.
\newblock How to work a crowd: Developing crowd capital through crowdsourcing.
\newblock {\em Business Horizons}, 58(1):77 -- 85.

\bibitem[Qi et~al., 2020]{qi2020stanza}
{\bf Qi, P.}, {\bf Zhang, Y.}, {\bf Zhang, Y.}, {\bf Bolton, J.}, \textbf{and}
  {\bf Manning, C.~D.} 2020.
\newblock Stanza: A {Python} natural language processing toolkit for many human
  languages.
\newblock In {\em Proceedings of the 58th Annual Meeting of the Association for
  Computational Linguistics: System Demonstrations}.

\bibitem[Ramisch et~al., 2018]{ramisch2018edition}
{\bf Ramisch, C.}, {\bf Cordeiro, S.}, {\bf Savary, A.}, {\bf Vincze, V.}, {\bf
  Mititelu, V.}, {\bf Bhatia, A.}, {\bf Buljan, M.}, {\bf Candito, M.}, {\bf
  Gantar, P.}, {\bf Giouli, V.}, \textbf{and} {\bf others} 2018.
\newblock Edition 1.1 of the {P}arseme shared task on automatic identification
  of verbal multiword expressions.

\bibitem[Rumshisky et~al., 2012]{rumshisky2012word}
{\bf Rumshisky, A.}, {\bf Botchan, N.}, {\bf Kushkuley, S.}, \textbf{and} {\bf
  Pustejovsky, J.} 2012.
\newblock Word sense inventories by non-experts.
\newblock In {\em Proceedings of the Eighth International Conference on
  Language Resources and Evaluation ({LREC}'12)}, pp. 4055--4059, Istanbul,
  Turkey. European Language Resources Association (ELRA).

\bibitem[Savary et~al., 2018]{Savarytv}
{\bf Savary, A.}, {\bf Candito, M.}, {\bf Mititelu, V.~B.}, {\bf Bejček, E.},
  {\bf Cap, F.}, {\bf Čéplö, S.}, {\bf Cordeiro, S.~R.}, {\bf Eryiğit, G.},
  {\bf Giouli, V.}, {\bf van Gompel, M.}, {\bf HaCohen-Kerner, Y.}, {\bf
  Kovalevskaitė, J.}, {\bf Krek, S.}, {\bf Liebeskind, C.}, {\bf Monti, J.},
  {\bf Escartín, C.~P.}, {\bf van~der Plas, L.}, {\bf QasemiZadeh, B.}, {\bf
  Ramisch, C.}, {\bf Sangati, F.}, {\bf Stoyanova, I.}, \textbf{and} {\bf
  Vincze, V.} 2018.
\newblock {PARSEME multilingual corpus of verbal multiword expressions}.
\newblock In {\bf Markantonatou, S.}, {\bf Ramisch, C.}, {\bf Savary, A.},
  \textbf{and} {\bf Vincze, V.}, editors, {\em {Multiword expressions at length
  and in depth{: Extended papers from the MWE 2017 workshop}}}, pp. 87--147.
  Language Science Press., Berlin.

\bibitem[Schneider et~al., 2014]{schneider2014comprehensive}
{\bf Schneider, N.}, {\bf Onuffer, S.}, {\bf Kazour, N.}, {\bf Danchik, E.},
  {\bf Mordowanec, M.~T.}, {\bf Conrad, H.}, \textbf{and} {\bf Smith, N.~A.}
  2014.
\newblock Comprehensive annotation of multiword expressions in a social web
  corpus.
\newblock In {\em Proceedings of the Ninth International Conference on Language
  Resources and Evaluation ({LREC}'14)}, pp. 455--461, Reykjavik, Iceland.
  European Language Resources Association (ELRA).

\bibitem[Siyanova-Chanturia, 2017]{teachingMWE2017}
{\bf Siyanova-Chanturia, A.} 2017.
\newblock Researching the teaching and learning of multi-word expressions.
\newblock {\em Language Teaching Research}, 21(3):289--297.

\bibitem[Snow et~al., 2008]{snow2008cheap}
{\bf Snow, R.}, {\bf O’connor, B.}, {\bf Jurafsky, D.}, \textbf{and} {\bf Ng,
  A.~Y.} 2008.
\newblock Cheap and fast--but is it good? evaluating non-expert annotations for
  natural language tasks.
\newblock In {\em Proceedings of the 2008 conference on EMNLP}, pp. 254--263.

\bibitem[Sprenger et~al., 2006]{sprenger2006lexical}
{\bf Sprenger, S.~A.}, {\bf Levelt, W.~J.}, \textbf{and} {\bf Kempen, G.} 2006.
\newblock Lexical access during the production of idiomatic phrases.
\newblock {\em Journal of memory and language}, 54(2):161--184.

\bibitem[Vasiljevic, 2015]{vasiljevic2015teaching}
{\bf Vasiljevic, Z.} 2015.
\newblock Teaching and learning idioms in l2: From theory to practice.
\newblock {\em Mextesol Journal}, 39(4):1--24.

\bibitem[Vincze et~al., 2011]{Vincze11}
{\bf Vincze, V.}, {\bf Nagy, I.}, \textbf{and} {\bf Berend, G.} 2011.
\newblock Multiword expressions and named entities in the {W}iki50 corpus.
\newblock In {\em Proc. of RANLP 2011}, pp. 289--295, Hissar.

\bibitem[Von~Ahn, 2006]{von2006games}
{\bf Von~Ahn, L.} 2006.
\newblock Games with a purpose.
\newblock {\em Computer}, 39(6):92--94.

\bibitem[Von~Ahn and Dabbish, 2004]{von2004labeling}
{\bf Von~Ahn, L.} \textbf{and} {\bf Dabbish, L.} 2004.
\newblock Labeling images with a computer game.
\newblock In {\em Proceedings of the SIGCHI conference on Human factors in
  computing systems}, pp. 319--326.

\bibitem[Von~Ahn et~al., 2006]{von2006verbosity}
{\bf Von~Ahn, L.}, {\bf Kedia, M.}, \textbf{and} {\bf Blum, M.} 2006.
\newblock Verbosity: a game for collecting common-sense facts.
\newblock In {\em Proceedings of the SIGCHI conference on Human Factors in
  computing systems}, pp. 75--78.

\end{thebibliography}
\label{lastpage}

\newpage 
\section*{Appendix}
\label{Sec:Appendix}

As stated in the introduction, deducting well-defined rules to express an idiom is usually a challenging task. Below we provide an example idiom from a language other than English\footnote{The example idiom is a Turkish idiom and is presented as ``(birinin) başının etini yemek'' in this language dictionary.}, with an explanation of its meaning and usage patterns in English (for a second language learner who speaks English).\\
 
Example idiom: $\ll$eat someone's head's meat$\gg$\\ ``annoying someone by talking too much'' as ``to nag at'' \\
Rule\#1: \textit{someone's} may be replaced with one of the possessive pronouns (e.g, my, your, his) or any noun taking a possessive suffix -’s (i.e. the genitive suffix in the target language). \\
Rule\#2: \textit{someone's} may be omitted since the target language is a pro-drop language and the word \textit{head} also takes possessive suffixes which also carry the person agreement information thus \textit{someone} is pragmatically or grammatically inferable. \\
Rule\#3: the verb \textit{eat} may be inflected\\
Rule\#4: since this language is an MRL and pro-drop language, the inflected verb will also carry the person agreement information thus the subject information coming with the verb (or either separately) should be different than \textit{someone}, i.e. reflexive usage is generally not welcome; ``eating own's head's meat''.\\

As one may notice, although it could be possible to define rules, they are both hard to deduct (e.g. for teachers or lexicographers) and hard to understand (for language learners: humans or computers). Language learners will still need usage examples both to understand the usage patterns and to practice. Additionally, 
for being able to define such rules even teachers or lexicographers should investigate many usage samples or come up with new ones.


\pagestyle{empty}
\begin{landscape}
\begin{table}[!htp]
\caption{Idioms of the TrP1 (first 16 rows) \& TrP2 (last 16 rows) -- (id.:idiomatic samples, nonid.:nonidiomatic samples, \rightthumbsdown:dislikes)}
\label{tab:TRrun}
\begin{center}
\vspace{-0.5cm}
\begin{adjustbox}{width=1.3\textwidth}
\begin{tabular}{rlllcccccc}
\hline
Day &
  Idiom &
  Literal Meaning &
  Idiomatic Meaning &
  \multicolumn{3}{l}{\begin{tabular}[c]{@{}l@{}}\# of Collected  Samples\\ Id. \hspace{0.3cm}   Nonid. \hspace{0.3cm}   Total\end{tabular}} &
  \multicolumn{1}{c}{\begin{tabular}[c]{@{}c@{}}\# of \\ Parseme  Id.\end{tabular}} &
  \multicolumn{1}{c}{\begin{tabular}[c]{@{}c@{}}Avg. \# of\\ Rewiews\end{tabular}} &
  \begin{tabular}[c]{@{}r@{}}\% of \\ \rightthumbsdown\end{tabular} \\ \hline
1  & hesap vermek     & bill - to give                    & to explain the reason for any behavior   & 198 & 212 & 410 & 5   & 4.6 & 7  \\
2  & altını çizmek    & to underline                      & to emphasize                             & 83  & 83  & 166 & 5   & 4.8 & 8  \\
3  & yer vermek       & place - to give                   & to emphasize the importance of sth       & 87  & 73  & 160 & 23  & 3.9 & 5  \\
4  & ayvayı yemek     & to eat a quince                   & to get in a bad situation                & 87  & 92  & 179 & 0   & 5.5 & 4  \\
5  & rol oynamak      & to act                            & to play an important role in sth         & 45  & 43  & 88  & 10  & 5.2 & 6  \\
6  & üzerinde durmak  & on top - to stand                 & to emphasize                             & 69  & 51  & 120 & 10  & 4.5 & 4  \\
7  & ağırlık vermek   & weight - to give                  & to emphasize                             & 49  & 48  & 97  & 8   & 3.8 & 5  \\
8  & yer almak        & place - to take/buy               & to occur                                 & 69  & 49  & 118 & 132 & 3.5 & 8  \\
9  & kolları sıvamak  & arms - to roll up                 & to get ready to do sth difficult         & 34  & 36  & 70  & 7   & 3.7 & 8  \\
10 & öne sürmek       & to put/drive to the front         & to suggest                               & 38  & 24  & 62  & 48  & 3.3 & 8  \\
11 & ortaya koymak    & to the middle - to put            & to introduce/to put forward              & 59  & 53  & 112 & 43  & 2.8 & 25 \\
12 & iz bırakmak      & a mark - to leave                 & to place in one's mind                   & 40  & 48  & 88  & 2   & 2.8 & 18 \\
13 & ele almak        & to the hand - to take             & to handle                                & 37  & 29  & 66  & 39  & 1.5 & 3  \\
14 & yol açmak        & a road - to open                  & to cause                                 & 21  & 20  & 41  & 38  & 2.2 & 6  \\
15 & meydana gelmek   & to the center - to come           & to happen                                & 11  & 18  & 29  & 29  & 2.2 & 2  \\
16 & karşı çıkmak     & in front of - to climb/to step up & to oppose                                & 14  & 7   & 21  & 27  & 1.2 & 8  \\ \hline
1  & içi erimek       & its inside - to melt              & to worry / to be upset                   & 121 & 149 & 270 & 0   & 5.5 & 19 \\
2  & el açmak         & hand - to open                    & to beg                                   & 206 & 211 & 417 & 0   & 3.0 & 17 \\ 
3  & zaman kazanmak   & time - to earn                    & to save time                             & 155 & 193 & 348 & 0   & 3.1 & 5  \\
4  & defterden silmek & to erase from notebook            & to forget someone                        & 97  & 99  & 196 & 1   & 3.9 & 28 \\
5  & nabzını tutmak   & to hold pulse                     & to measure intension                     & 58  & 55  & 113 & 2   & 4.0 & 7  \\
6  & basamak yapmak   & step - to make                    & to exploit someone                       & 79  & 95  & 174 & 0   & 4.2 & 26 \\
7  & başa geçmek      & to the head - to pass             & to govern                                & 68  & 58  & 126 & 1  & 3.5 & 2  \\
8  & sıkı durmak      & to stay/to look tight             & to stay strong or to be ready            & 201 & 173 & 374 & 0   & 2.7 & 36 \\
9  & üste çıkmak      & to the top - to climb/to step up  & to blame others even though being guilty & 85  & 176 & 261 & 0   & 2.9 & 26 \\
10 & sayıp dökmek     & to count and to pour              & to tell everything                       & 74  & 133 & 207 & 0   & 3.3 & 10 \\
11 & üstünden atmak   & from over to throw                & to get rid of                            & 65  & 80  & 145 & 0   & 4.7 & 18 \\
12 & zaman öldürmek   & time - to kill                    & to waste time                            & 123 & 127 & 250 & 1   & 3.7 & 14 \\
13 & üstüne almak     & onto - to take                    & to undertake                             & 113 & 300 & 413 & 1   & 4.0 & 3  \\
14 & parmak basmak    & finger - to press                 & to attract attention on sth              & 54  & 89  & 143 & 1   & 4.3 & 6  \\
15 & meydana gelmek   & to come to the center             & to happen                                & 103 & 92  & 195 & 29  & 3.6 & 29 \\
16 & karşı çıkmak     & to climb/step up opposite      & to oppose                                & 167 & 352 & 519 & 27  & 4.1 & 6  \\ \hline
\end{tabular}

\end{adjustbox}
\end{center}

\end{table}

\end{landscape}
\pagestyle{plain}

\pagestyle{empty}
\begin{landscape}
\begin{table}[!htp]
\caption{Idioms of the ItP1 (first 16 rows) \& ItP2 (last 16 rows) -- (id.:idiomatic samples, nonid.:nonidiomatic samples, \rightthumbsdown:dislikes)}
\label{tab:ITRun}
\begin{center}
\vspace{-0.5cm}
\begin{adjustbox}{width=1.4\textwidth}
\begin{tabular}{rlllcccccc}
\hline
Day &
  Idiom &
  Literal Meaning &
  Idiomatic Meaning &
  \multicolumn{3}{l}{\begin{tabular}[c]{@{}l@{}}\# of Collected  Samples\\ Id. \hspace{0.3cm}   Nonid. \hspace{0.3cm}   Total\end{tabular}} &
  \multicolumn{1}{c}{\begin{tabular}[c]{@{}c@{}}\# of \\ Parseme  Id.\end{tabular}} &
  \multicolumn{1}{c}{\begin{tabular}[c]{@{}c@{}}Avg. \# of\\ Rewiews\end{tabular}} &
  \begin{tabular}[c]{@{}r@{}}\% of \\ \rightthumbsdown\end{tabular} \\ \hline
1 & gettare la spugna & to throw the sponge & to throw in the towel & 470 & 302 & 772 & 0 & 3.6 & 29 \\
2 & coltivare il proprio orto & to cultivate one's vegetable garden & to care only about one's problems & 61 & 79 & 140 & 0 & 3.6 & 42 \\
3 & buttare giu & to throw down & to swallow, to overthrow, to push over & 43 & 41 & 84 & 0 & 3.6 & 32 \\
4 & mettere dentro & to put inside & to put in jail & 75 & 130 & 205 & 0 & 3.1 & 40 \\
5 & abbaiare alla luna & to bark to the moon & to bark at the moon, to swear & 58 & 23 & 81 & 0 & 3.4 & 33 \\
6 & acchiappare farfalle & to catch butterflies & to do useless things & 21 & 8 & 29 & 0 & 1.9 & 35 \\
7 & ingoiare una pillola & to swallow a pill & to subject oneself to something unpleasant & 7 & 3 & 10 & 0 & 1.7 & 6 \\
8 & ammainare le vele & to furl the sails & to abandon, to surrender & 10 & 1 & 11 & 0 & 0.4 & 25 \\
9 & andare a gonfie vele & to go with inflated sails & to be successful & 23 & 4 & 27 & 0 & 3.3 & 6 \\
10 & andare in barca & to go in boat & to break down & 14 & 8 & 22 & 0 & 2.1 & 2 \\
11 & aprire gli occhi & to open the eyes & to awaken, to realize & 2 & 1 & 3 & 2 & 0.0 & 0 \\
12 & attaccare bottone & to attach button & to chat up, to talk endlessly & 3 & 3 & 6 & 0 & 0.0 & 0 \\
13 & avere la coda di paglia & to have the tail of straw & to feel guilty & 27 & 10 & 37 & 0 & 1.8 & 1 \\
14 & avere la corda al collo & to have the rope at the neck & to not have control & 9 & 0 & 9 & 0 & 0.4 & 25 \\
15 & avere le mani lunghe & to have the hands long & to steal & 4 & 2 & 6 & 0 & 0.7 & 0 \\
16 & avere birra in corpo & to have beer in body & to have strength & 1 & 0 & 1 & 0 & 0.0 & 0 \\ \hline
1 & avere il becco lungo & to have the beak long & to speak outright & 185 & 98 & 283 & 0 & 3.9 & 19 \\
2 & avere il mestolo in mano & to have the ladle in hand & to rule despotically & 277 & 421 & 698 & 0 & 1.6 & 24 \\
3 & prendere con le pinze & to take with the pincers & to take it with a pinch of salt & 409 & 394 & 803 & 1 & 0.8 & 32 \\
4 & raggiungere il bersaglio & to reachthe target & to reach the objective & 285 & 89 & 374 & 0 & 1.5 & 23 \\
5 & buttare al vento & to throw to the wind & to fritter away & 188 & 162 & 350 & 0 & 0.7 & 35 \\
6 & brancolare nel buio & to grope in the dark & to grope in the dark & 334 & 246 & 580 & 0 & 1.0 & 49 \\
7 & attaccare bottone & to attach button & to talk endlessly & 53 & 68 & 121 & 0 & 2.7 & 8 \\
8 & avere le batterie scariche & to have the batteries dead & to be exhausted & 156 & 101 & 257 & 0 & 2.0 & 34 \\
9 & aprire gli occhi & to open the eyes & to awaken, to realize & 143 & 132 & 275 & 2 & 1.0 & 22 \\
10 & portare a casa & to take home & to earn & 113 & 75 & 188 & 2 & 2.6 & 17 \\
11 & tirare su & to pull up & to raise & 135 & 64 & 199 & 0 & 1.3 & 44 \\
12 & essere tra i piedi & to be among the feet & to  get in sb's way & 152 & 32 & 184 & 2 & 2.2 & 17 \\
13 & dare corda & to give cord & to give sb a free hand & 339 & 130 & 469 & 0 & 1.8 & 35 \\
14 & mandare a casa & to send at house & to send away, to dispatch, to kick out & 102 & 137 & 239 & 2 & 2.4 & 6 \\
15 & avere le mani lunghe & to have the hand long & to steal & 115 & 32 & 147 & 0 & 2.6 & 15 \\
16 & avere birra in corpo & to have beer in body & to have strength & 83 & 35 & 118 & 0 & 4.0 & 6 \\ 
\hline
\end{tabular}

\end{adjustbox}
\end{center}

\end{table}
\end{landscape}
\pagestyle{plain}

\end{document}